\documentclass[sigconf,nonacm]{aamas}

\usepackage{balance}
\usepackage{multirow}
\usepackage{amsmath}
\usepackage{svg}
\setcopyright{none}
\hypersetup{
  pdftitle={LifelongCrossNav: Persistent 3D Semantic Memory for Cross-Floor Multi-Object Navigation},
  pdfauthor={Zehui Li, Zihao Sun, Jiawei Xu, Zheqi He, Xiaoqiang Zhang, Jing-Shu Zheng, Lu Liu, Dahui Gao, Xiuwan Chen},
  pdfsubject={Research preprint}
}

\title[LifelongCrossNav]{LifelongCrossNav: Persistent 3D Semantic Memory for
Cross-Floor Multi-Object Navigation}

\author[Li et al.]{Zehui Li\textsuperscript{1,*}}
\affiliation{\institution{Peking University}\city{Beijing}\country{China}}
\email{zehui.li@stu.pku.edu.cn}
\author{Zihao Sun\textsuperscript{1,*}}
\affiliation{\institution{Peking University}\city{Beijing}\country{China}}
\author{Jiawei Xu\textsuperscript{1}}
\affiliation{\institution{Peking University}\city{Beijing}\country{China}}
\author{Zheqi He\textsuperscript{2,\textdagger}}
\affiliation{\institution{Beijing Academy of Artificial Intelligence}\city{Beijing}\country{China}}
\email{zqhe@baai.ac.cn}
\author{Xiaoqiang Zhang\textsuperscript{1,\textdagger}}
\affiliation{\institution{Peking University}\city{Beijing}\country{China}}
\author{Jing-Shu Zheng\textsuperscript{2}}
\affiliation{\institution{Beijing Academy of Artificial Intelligence}\city{Beijing}\country{China}}
\author{Lu Liu\textsuperscript{2}}
\affiliation{\institution{Beijing Academy of Artificial Intelligence}\city{Beijing}\country{China}}
\author{Dahui Gao\textsuperscript{2}}
\affiliation{\institution{Beijing Academy of Artificial Intelligence}\city{Beijing}\country{China}}
\author{Xiuwan Chen\textsuperscript{1}}
\affiliation{\institution{Peking University}\city{Beijing}\country{China}}
\thanks{\textsuperscript{*}Equal contribution. \textsuperscript{\textdagger}Corresponding authors.}

\begin{document}
\begin{abstract}
Object-goal navigation has made substantial progress in
semantic perception and exploration, yet persistent memory
for multi-object navigation and cross-floor navigation are
still commonly addressed separately. We present
\textbf{LifelongCrossNav}, a framework for sequential
multi-object ObjectNav in unknown multi-floor indoor
environments. Within each episode, the agent receives an
ordered sequence of object-goal queries while continuously
maintaining a shared sparse 3D semantic voxel memory. This
memory incrementally accumulates geometric structure,
traversability states, and vision-language features, allowing
subsequent object-goal queries to retrieve previously acquired
scene information without rebuilding the map. To support
persistent search across floors, LifelongCrossNav combines
support-aware 3D traversability mapping, stair-specific
perception, and direction-aware stair traversal. A unified
navigation policy coordinates same-floor frontier exploration,
live and historical point-of-interest retrieval, stair
navigation, and target-object search and approach. We further
introduce \textbf{HM3D-MFMON}, a benchmark for sequential
\textbf{M}ulti-\textbf{F}loor \textbf{M}ulti-\textbf{O}bject \textbf{N}avigation built on HM3D scenes, including a dedicated subset in which completing the full sequence of object-goal subtasks requires at least one floor transition.
Experimental results show that LifelongCrossNav consistently outperforms a representative planar persistent semantic-map baseline on HM3D-MFMON,
demonstrating that persistent 3D semantic memory and cross-floor traversability modeling effectively support sequential
multi-object navigation in multi-floor environments.

\end{abstract}
\maketitle
\pagestyle{plain}
\thispagestyle{plain}
\hypersetup{pdfauthor={Zehui Li, Zihao Sun, Jiawei Xu, Zheqi He, Xiaoqiang Zhang, Jing-Shu Zheng, Lu Liu, Dahui Gao, Xiuwan Chen}, pdfcreator={LaTeX}, pdfsubject={Research preprint}, pdfkeywords={embodied navigation, cross-floor navigation, semantic memory}}
\section{Introduction}

\begin{figure}[t]
    \centering
    \includegraphics[width=0.9\columnwidth]{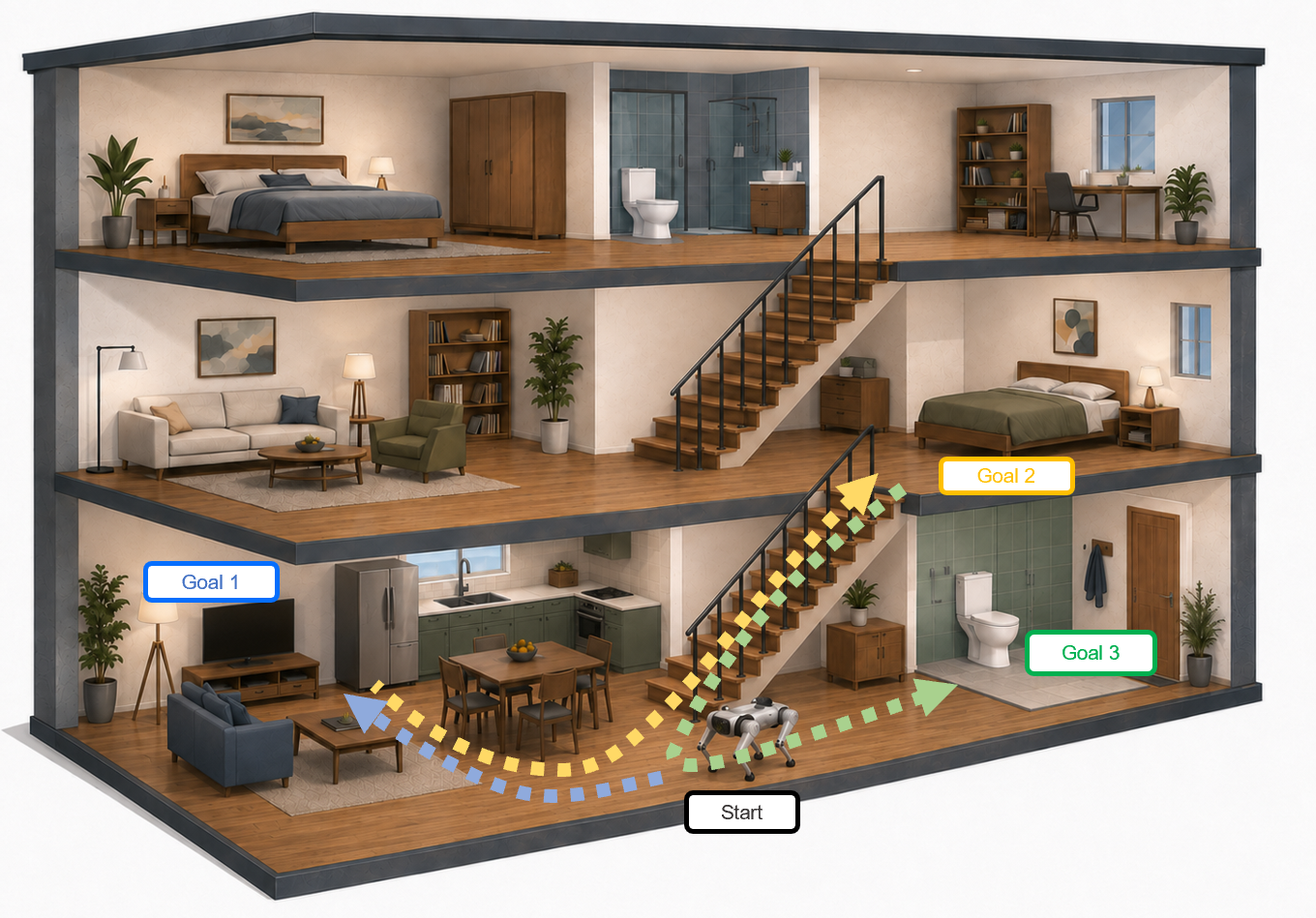}
    \caption{
    Conceptual illustration of sequential multi-object ObjectNav
in a multi-floor environment. The agent navigates from Start to Goal~1 (\emph{TV}), Goal~2 (\emph{bed}), and Goal~3 (\emph{toilet}) in order. The colored trajectories denote consecutive object-goal subtasks, including cross-floor traversal through the staircase.
    }
    \label{fig:overall_illustration}
    \Description{
Conceptual multi-floor indoor environment containing an agent
start position and three sequential object goals: a television,
a bed, and a toilet. Colored trajectories illustrate consecutive
navigation subtasks, including traversal through a staircase.
}
\end{figure}

Object-Goal Navigation (ObjectNav) requires an embodied
agent to explore an unseen environment and navigate to an
instance of a specified object category. Recent advances in
open-vocabulary perception, semantic mapping, and
frontier-based exploration have substantially improved target
search in unseen scenes
\cite{duan2022embodiedsurvey,sun2025objectnavsurvey,
liu2025embodiednavigation}. However, most existing studies
remain centered on single-object navigation in planar or
single-floor environments, whereas real indoor agents may
need to search for multiple objects in sequence and
move between floors, as illustrated in Fig.~\ref{fig:overall_illustration}.

Existing research largely addresses these requirements
separately. Cross-floor methods model stairs, floor
transitions, or floor-level reasoning, but generally consider
only one object goal
\cite{gong2026ascent,zheng2026travexplorer}. Multi-object
navigation instead retains environmental information across
a sequence of object goals
\cite{wani2020multion,busch2025onemap}. In this setting, the
agent receives only the current object goal, and the next one
is revealed after the current object-goal subtask is completed.
Following prior lifelong embodied-navigation settings
\cite{khanna2024goat}, we use \emph{lifelong} to denote
persistent operation across sequential object-goal queries in
the same environment: the scene and accumulated memory are
retained as new goals are issued. This setting concerns
cross-goal memory reuse.
Nevertheless, existing multi-object methods primarily rely on
planar representations or avoid goal sequences that require
stair traversal. Planar maps may collapse vertically
overlapping spaces, while single-object cross-floor methods
need not preserve queryable semantic observations for future
object goals. A unified solution must therefore support both
vertical traversability and persistent semantic memory.

To address this gap, we present
\textbf{LifelongCrossNav}, a framework for sequential
multi-object ObjectNav in unknown multi-floor indoor
environments. It maintains a shared sparse 3D voxel
representation that combines support-aware geometry with
goal-independent vision-language features. The stored
features can be re-queried when a new object goal is issued,
while a unified navigation policy coordinates same-floor
exploration, stair traversal, semantic retrieval, and final
target-object approach through mode-aware 3D planning.
We further introduce \textbf{HM3D-MFMON}, a benchmark for
sequential multi-object navigation in multi-floor HM3D
scenes. To handle multiple valid target object instances and
agent-dependent subtask starting positions, we adopt a
post-hoc stage-wise shortest-path protocol and conditional
metrics that evaluate agents reaching each successive
object-goal subtask. 
Our contributions are threefold:
\begin{itemize}
    \item We introduce HM3D-MFMON, comprising 927
    three-goal episodes from 36 multi-floor HM3D scenes,
    including 288 Cross-Floor-Required episodes, together
    with a post-hoc evaluation protocol for multi-instance
    sequential navigation.

    \item We propose LifelongCrossNav, which unifies
    support-aware 3D voxel mapping, persistent
    vision-language memory, stair-aware navigation, and
    historical semantic retrieval in a closed-loop system.

    \item  Experiments demonstrate improved multi-object and
    cross-floor navigation over a persistent planar semantic memory baseline.
    A cross-target memory ablation further demonstrates
    that persistent memory reuse improves navigation efficiency
    across sequential object goals.
\end{itemize}
\section{Related Work}

\subsection{Open-Vocabulary Semantic Navigation}

Early ObjectNav methods combined semantic perception with
explicit spatial memory. Goal-Oriented Semantic Exploration,
for example, projects object detections into a semantic map
and uses frontier-based exploration for long-horizon
navigation
\cite{batra2020objectnavrevisited,chaplot2020objectgoal}.
More recent methods use pretrained vision-language models
to support zero-shot or open-vocabulary target search.
ZSON aligns object goals with visual observations through
multimodal goal embeddings, while VLFM scores exploration
frontiers using vision-language relevance
\cite{majumdar2022zson,yokoyama2024vlfm}.
OpenFMNav and SG-Nav further incorporate foundation models
or online 3D scene graphs for semantic reasoning
\cite{kuang2024openfmnav,yin2024sgnav}. These methods
improve generalization to unseen object categories, but they
primarily evaluate single-object navigation and do not focus
on retaining semantic memory across successive object goals.

\subsection{Multi-Object Navigation and Persistent Memory}

Multi-object navigation extends ObjectNav from one object
goal to an ordered sequence of object goals within the same
environment. MultiON introduced this setting to evaluate
semantic mapping and memory over successive target searches
\cite{wani2020multion}. GOAT-Bench further studies lifelong
navigation with object-category, image, and language goals,
emphasizing the reuse of experience without resetting the
environment \cite{khanna2024goat}. OneMap addresses
zero-shot multi-object navigation by maintaining a reusable
open-vocabulary feature map that can be queried again when
a new object goal is issued \cite{busch2025onemap}.
These works demonstrate that persistent scene memory can
reduce repeated exploration and improve later-goal
efficiency. However, their navigation representations are
primarily planar or bird's-eye-view maps and do not explicitly
model stairs, vertically overlapping spaces, or traversable
connections between floors.

\subsection{Cross-Floor Navigation and 3D Representation}

Cross-floor ObjectNav considers environments in which the
agent and a target object instance may lie on different
floors. MFNP uses multimodal reasoning and floor-transition
policies for multi-floor target search
\cite{zhang2025mfnp}, while ASCENT combines a multi-floor
spatial abstraction with stair-aware, coarse-to-fine
exploration \cite{gong2026ascent}. TravExplorer instead
maintains a unified traversability-aware 3D representation of
floors, stairs, and landings, allowing cross-floor paths to be
planned directly through connected support surfaces
\cite{zheng2026travexplorer}. Related 3D semantic mapping
work, such as BeliefMapNav, shows that voxel maps can also
organize object-location priors, online observations, and
semantic uncertainty \cite{zhou2025beliefmapnav}.
These cross-floor approaches primarily address single-goal
episodes, while persistent multi-object methods do not explicitly
model vertically connected traversable space. LifelongCrossNav
targets the intersection of these two settings by retaining
queryable 3D scene memory across sequential object goals while
supporting explicit cross-floor navigation.

\section{Method}

\subsection{Task Formulation}

Each episode is defined by an ordered sequence of $K$ object
goals $(g_1,\ldots,g_K)$. Following sequential multi-object
navigation, the agent receives only the current object goal
$g_k$; the next object goal $g_{k+1}$ is revealed and
activated only after the current object-goal subtask has been
successfully completed \cite{wani2020multion}.
At each time step $t$, the agent receives an egocentric RGB
observation $I_t$, a depth observation $D_t$, the 6-DoF
camera pose $T_t$, and the current object goal $g_k$. The
discrete action space contains \texttt{MoveForward},
\texttt{TurnLeft}, \texttt{TurnRight}, \texttt{LookUp},
\texttt{LookDown}, and \texttt{Stop}. \texttt{MoveForward}
advances the agent by $0.25$\,m, while turning and camera
pitch actions change the corresponding orientation by
$30^{\circ}$. A subtask is successful when the agent issues
\texttt{Stop} within the success threshold of a valid target
object instance.

\begin{figure*}[t]
    \centering
    \includegraphics[width=0.9\textwidth]{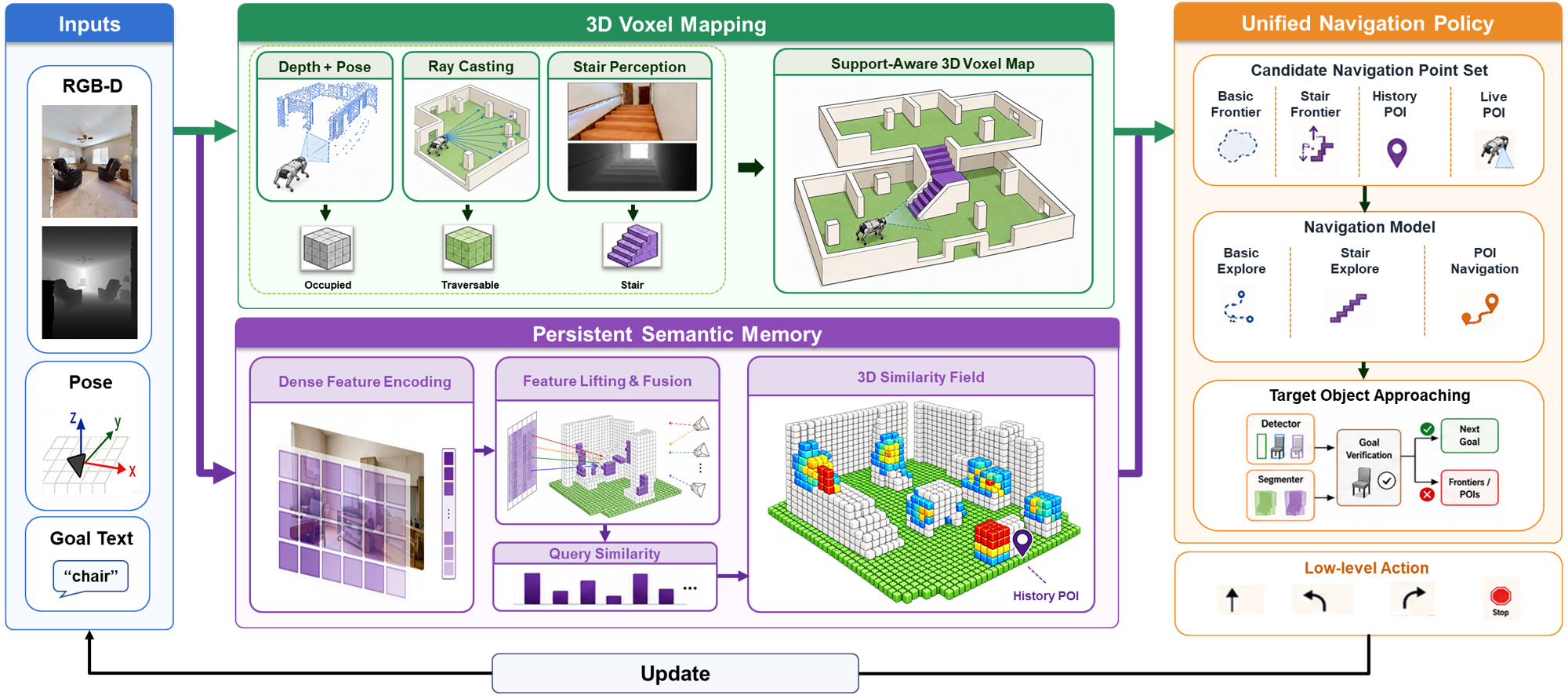}
    \caption{
    Overview of LifelongCrossNav. Given RGB-D observations,
    the agent pose, and the current goal text, the system jointly
    updates a support-aware 3D voxel map and persistent 3D
    semantic memory. The geometric branch constructs multi-floor
    traversability from observed surfaces, ray-cast free space,
    and semantic-geometric stair evidence, while the semantic
    branch lifts and fuses goal-independent vision-language
    features into 3D surface voxels. The unified navigation policy
    selects among Basic Frontiers, Stair Frontiers, History POIs,
    and Live POIs and performs mode-aware 3D planning for basic
    exploration, stair exploration, POI navigation, and final
    object approach. After the current object goal is completed,
    the accumulated environmental memory is retained and the next
    object goal is activated.
    }
    \label{fig:framework}
    \Description{
Overview of LifelongCrossNav. RGB-D observations, camera
pose, and goal text are processed by support-aware 3D voxel
mapping and persistent semantic-memory branches. Their outputs
are used by a unified navigation policy for frontier exploration,
stair traversal, semantic-POI navigation, and final target approach.
}
\end{figure*}

Within an episode, LifelongCrossNav retains the 3D geometric
map, stair states, and goal-independent vision-language voxel
features across object goals. When a new object goal is
issued, its text embedding and query-conditioned similarity
field are recomputed, while the POIs, navigation path, and
goal-verification states associated with the previous object
goal are reinitialized. An overview of LifelongCrossNav is
shown in Fig.~\ref{fig:framework}.

\subsection{Support-Aware 3D Voxel Mapping}

LifelongCrossNav represents the environment using a sparse
3D voxel map. In contrast to planar projection, this
representation preserves height and distinguishes vertically
overlapping rooms, corridors, landings, and staircases.
Inspired by traversability-aware ray-casting approaches
\cite{zheng2026travexplorer}, the map combines RGB-D
geometry, vertical support relationships, and stair-semantic
evidence to represent executable connections across floors.

\noindent\textbf{RGB-D Projection and Sparse Voxelization.}
Valid depth pixels are back-projected into the camera
coordinate system, transformed into the world frame using
the 6-DoF camera pose, and quantized into sparse voxels.
Depth-ray endpoints provide observed surface evidence,
whereas intermediate ray locations provide free-space
evidence. Only observed voxels and locally inferred
navigation states are stored, allowing the map to grow
incrementally with the explored space.

\noindent\textbf{Support-Aware Voxel Types.}
The navigation map abstracts the observed space into four
functional voxel types: \emph{Occupied}, \emph{Traversable},
\emph{Stair}, and \emph{Unsupported}. Occupied voxels
represent observed surfaces, including walls, furniture,
floors, and stair candidates that have not yet been confirmed.
For each ray-observed free-space voxel, the system searches
downward within a local vertical range. Free space with
reliable support is classified as Traversable, whereas free
space without observed support is classified as Unsupported
and provides geometric evidence for potential downward
transitions.

Stair voxels represent stair surfaces confirmed jointly by
semantic and geometric observations. SegFormer-B2
\cite{xie2021segformer} extracts a pixel-level stair mask,
which is lifted into 3D using depth and camera pose and
verified through local height variation, step trends, spatial
continuity, and support relationships. Stair voxels are updated
only after the system enters \emph{Stair Explore}, where they
form an explicit traversable connection between floors.

\subsection{Persistent 3D Semantic Memory}

LifelongCrossNav stores goal-independent vision-language features
in the same sparse 3D coordinate system and re-queries them
whenever the active object goal changes.

\noindent\textbf{Dense Vision-Language Feature Encoding.}
We adopt SED-based dense vision-language encoding
\cite{xie2024sed}. Given the current RGB observation, the
encoder produces a $24\times24\times768$ spatial feature
map aligned with the CLIP text-embedding space. The feature
map is bilinearly upsampled to the depth resolution, and each
valid depth pixel is associated with a local-contextual visual
feature.

\noindent\textbf{Feature Lifting and Cumulative Fusion.}
Pixel features are lifted to the 3D surface voxels reached by
their corresponding depth rays. Features are assigned only
to observed surfaces and are not propagated through
free-space voxels. When multiple pixels in the same frame
are quantized into voxel $\mathbf{v}$, their features are
aggregated using observation-quality weights:
\begin{equation}
    \bar{\mathbf{f}}_{\mathbf{v}}^{(t)}
    =
    \frac{
        \sum_{p\in\mathcal{P}_{\mathbf{v}}}
        q_p \mathbf{f}_p
    }{
        \sum_{p\in\mathcal{P}_{\mathbf{v}}}q_p+\epsilon
    },
    \label{eq:intra_frame_fusion}
\end{equation}
where $\mathcal{P}_{\mathbf{v}}$ is the set of pixels assigned
to voxel $\mathbf{v}$, $\mathbf{f}_p$ is the corresponding
visual feature, and $q_p$ reflects the observation quality.

Let $\mathbf{F}_{\mathbf{v}}^{(t-1)}$ and
$C_{\mathbf{v}}^{(t-1)}$ denote the stored feature and
accumulated confidence, and let $c_{\mathbf{v}}^{(t)}$ denote
the confidence of the current per-frame observation.
Multi-view observations are fused using a cumulative weighted
average:
\begin{equation}
\begin{aligned}
    \mathbf{F}_{\mathbf{v}}^{(t)}
    &=
    \frac{
        C_{\mathbf{v}}^{(t-1)}
        \mathbf{F}_{\mathbf{v}}^{(t-1)}
        +
        c_{\mathbf{v}}^{(t)}
        \bar{\mathbf{f}}_{\mathbf{v}}^{(t)}
    }{
        C_{\mathbf{v}}^{(t-1)}
        +
        c_{\mathbf{v}}^{(t)}
    }, \\
    C_{\mathbf{v}}^{(t)}
    &=
    C_{\mathbf{v}}^{(t-1)}
    +
    c_{\mathbf{v}}^{(t)} .
\end{aligned}
\label{eq:inter_frame_fusion}
\end{equation}

\noindent\textbf{Query-Conditioned 3D Retrieval.}
When an object goal is activated, its text embedding is
compared with the stored surface-voxel features using cosine
similarity. The resulting responses form a query-conditioned
3D similarity field without rebuilding the geometric or
semantic map. High-response historical regions are grouped
using 3D neighborhood clustering, after which the Top-$K$
candidates are retained as History POIs and assigned nearby
reachable navigation positions.

\subsection{Unified Navigation Policy}

The unified policy selects navigation targets from Basic
Frontiers, Stair Frontiers, History POIs, and Live POIs.
These candidates support three navigation modes:
\emph{Basic Explore}, \emph{Stair Explore}, and
\emph{POI Navigation}.

\begin{figure}[t]
    \centering
    \includegraphics[width=0.9\columnwidth]{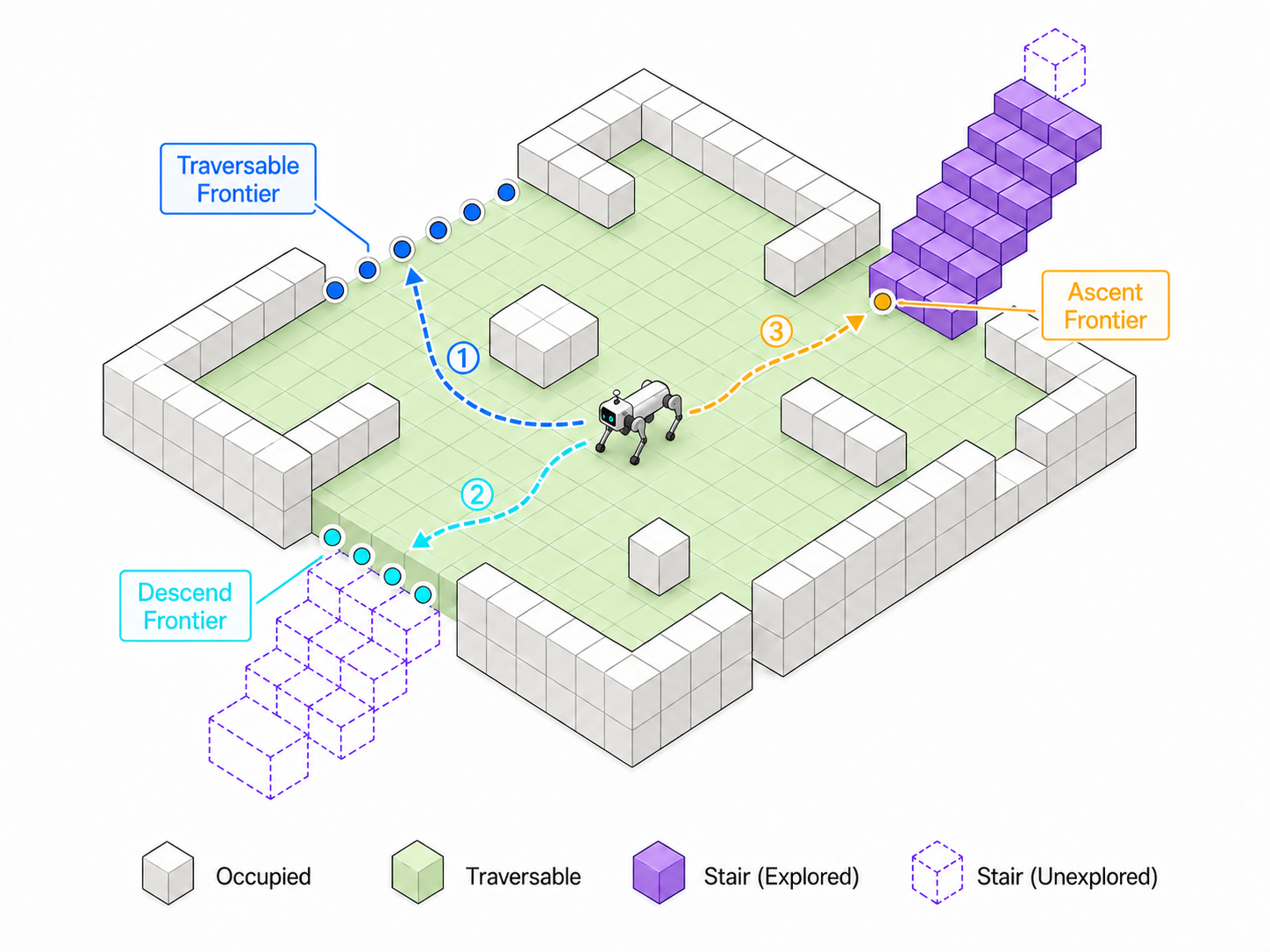}
    \caption{
    Basic Frontier types and their default priority:
    \textcircled{1} Traversable, \textcircled{2} Descend, and
    \textcircled{3} Ascent. The dashed staircase denotes
    unexplored cross-floor structure.
    }
    \label{fig:frontier}
    \Description{
Schematic illustration of three exploration-frontier types:
traversable frontiers toward unknown same-floor space,
descend frontiers near unsupported regions, and ascent
frontiers near candidate stair entrances.
}
\end{figure}

\noindent\textbf{Candidate Navigation Points.}
Basic Frontiers denote the three planar candidate types used
during ordinary exploration, as shown in Fig.~\ref{fig:frontier}. Traversable Frontiers are formed
at boundaries between Traversable voxels and unknown space.
Descend Frontiers are extracted near boundaries between
Traversable and Unsupported voxels and indicate potential
downward transitions. Ascent Frontiers are generated from
stair-semantic evidence, RGB-D step geometry, and
multi-frame consistency, with their navigation positions
remaining on the current exploration plane near a potential
ascending entrance.

Stair Frontiers are direction-aware advancing boundaries
extracted from the currently confirmed stair-voxel cluster.
Their heights change with the observed staircase, allowing
the agent to continue exploring upward or downward within
the stair structure.

History POIs are obtained by querying the persistent
vision-language voxel memory with the current goal text,
thereby supporting semantic retrieval beyond a fixed detector
label set. Live POIs are generated through a configurable
target-detection interface. We employ
D-FINE~\cite{peng2025dfine} for the six HM3D
ObjectNav target categories, which are covered by
MS-COCO, and can switch to YOLO-World
\cite{cheng2024yoloworld} for queries outside this category
set. MobileSAM~\cite{zhang2023mobilesam} extracts the detected-object mask, while
depth and pose recover its 3D location.

\noindent\textbf{Hierarchical Mode Switching and 3D Planning.}
Reliable current target evidence and reachable POIs are
prioritized over unexplored geometric candidates. When no
valid semantic candidate is available, Basic Explore first
visits Traversable Frontiers on the current floor. Descend and
Ascent Frontiers are considered only after no reachable
ordinary frontier remains. Once a stair entrance has been
verified, the policy activates Stair Explore and follows
Stair Frontiers until a new landing is reached. An active stair
session retains control so that stair traversal is not
interrupted by ordinary frontiers or unrelated semantic
candidates.

All modes use a common 3D A* planner but differ in
admissible voxel types and graph connectivity. Basic Explore
uses a strict 26-neighbor graph over the current traversable
region. POI Navigation uses the regular graph together with
previously established stair connectivity when the selected
POI lies on another floor. Stair Explore admits confirmed
Stair voxels and expands the endpoint neighborhood within a
fixed physical radius to connect staircase observations that
may be sparse in the voxel map.

\noindent\textbf{Final Target-Object Approach and Verification.}
Once stable target-object evidence is obtained, the detected
mask is projected into 3D and a nearby safe observation
position is selected. During the approach, the system updates
the detection and verifies the target object using detection
confidence, mask quality, visibility, viewing direction, 
3D distance, and SED similarity. 

The final SED check requires the mean of the top 20\%
text--region similarities within the object mask to reach
0.15. Rejected candidates are temporarily suppressed,
and the agent resumes navigation.
Reaching either a History POI or a Live POI
does not by itself complete the subtask; a current
target-object observation must still pass the verification
criteria. After successful verification, the next object goal
is activated while the geometric map, stair structure, and
goal-independent semantic memory are retained. If
verification fails, the system returns to candidate selection
and continues exploration.

\section{Experimental Setup}

\subsection{HM3D-MFMON Benchmark}

We construct HM3D-MFMON from HM3D v0.2 and its semantic
annotations~\cite{ramakrishnan2021hm3d,yadav2023hm3dsem},
selecting 36 scenes with realistic multi-floor layouts and
navigable stair connections. These scenes support cross-floor
navigation while remaining compatible with established
Habitat ObjectNav protocols.

Following the sequential task formulation of
MultiON~\cite{wani2020multion}, we generate 927 episodes,
each containing three sequentially issued object goals.
The agent receives only the current goal, with the next
revealed upon successful completion. The resulting 2,781
goal queries cover six categories: \texttt{chair} (19.24\%),
\texttt{bed} (16.90\%), \texttt{toilet} (17.33\%),
\texttt{plant} (13.88\%), \texttt{sofa} (18.66\%), and
\texttt{tv\_monitor} (13.99\%). All valid instances of
each target category are retained, and reaching any valid
instance completes the corresponding subtask.

By verifying target-instance distributions and navigable
connectivity, we identify 288 Cross-Floor-Required (CFR)
episodes whose complete object-goal sequences necessarily
require at least one floor transition.

\subsection{Post-hoc Multi-Object Evaluation Protocol}

\noindent\textbf{Post-hoc Stage-Wise Shortest-Path Evaluation.}
In standard single-object ObjectNav, the episode start and
target object category are fixed, allowing the shortest
geodesic distance to the nearest valid target instance to be
computed before navigation \cite{batra2020objectnavrevisited}.
This assumption does not directly extend to sequential
multi-object navigation with multiple valid instances. The
start of a later subtask depends on the target object instance
selected for the previous object goal and on the position from
which that subtask was completed.

\begin{figure}[t]
    \centering
    \includegraphics[width=0.9\columnwidth]{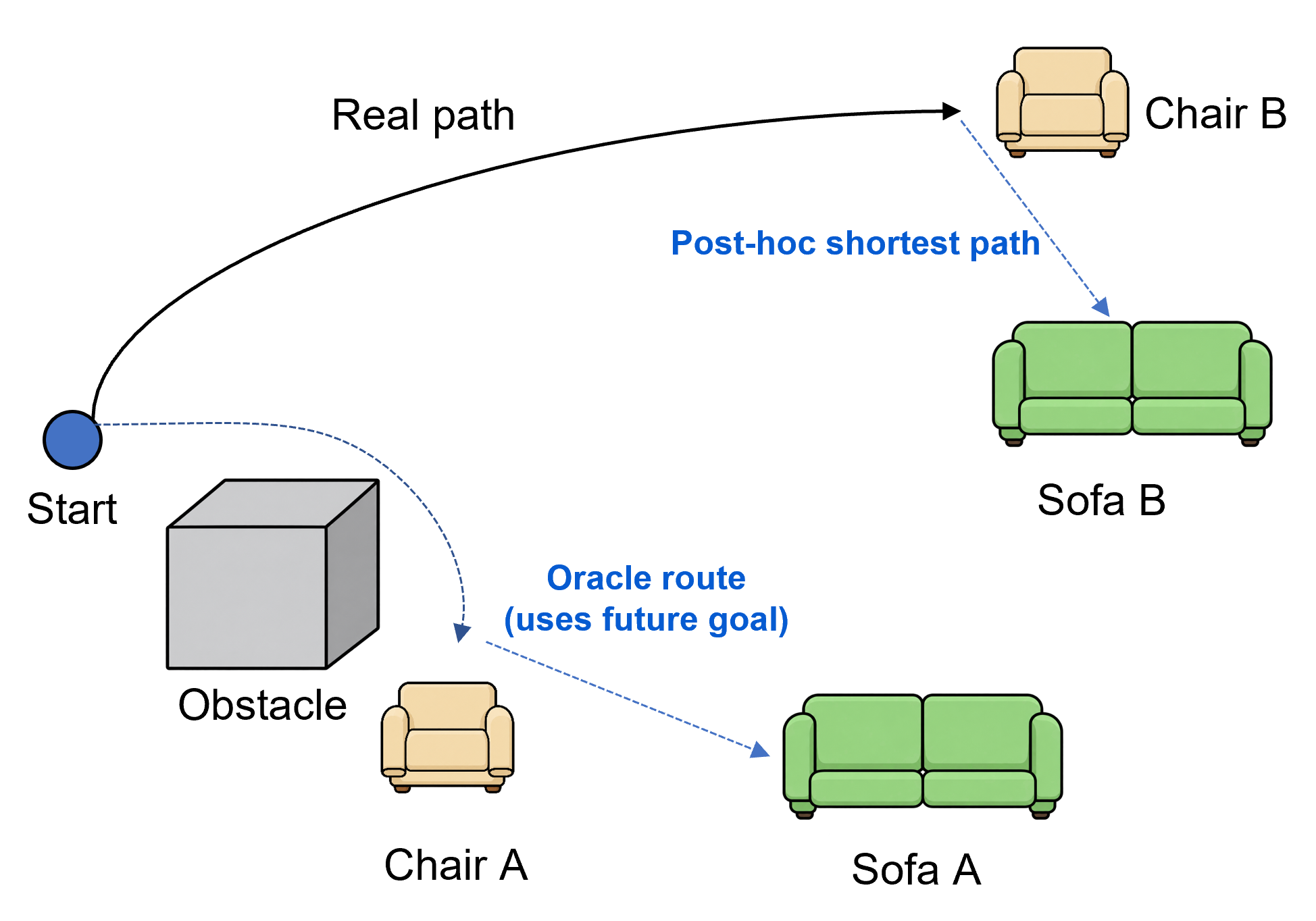}
    \caption{
    Post-hoc stage-wise shortest-path evaluation. A globally
    optimized route may depend on future object goals that are
    unavailable to the agent.
    }
    \label{fig:posthoc_protocol}
    \Description{
Illustration comparing post-hoc stage-wise shortest-path
evaluation with a globally optimized oracle route. The oracle
route can select the current target instance using knowledge of
a future goal, whereas the post-hoc route uses only the actual
start position and the currently active object goal.
}
\end{figure}

A globally optimized route over the complete object-goal
sequence could resolve this ambiguity, but such a route uses
future object goals that have not yet been issued. It may
therefore favor a non-nearest instance of the current target
category solely because that instance is closer to a future
goal, violating the causal information constraint of the
sequential task. Figure~\ref{fig:posthoc_protocol} illustrates
the difference between such an oracle route and the proposed
stage-wise evaluation.

For episode $i$, let
$\mathbf{x}^{\mathrm{start}}_{i,k}$ denote the actual agent
position when the $k$-th object goal $g_{i,k}$ is activated.
For the first subtask, this is the episode start; for later
subtasks, it is obtained from the executed trajectory after
the preceding object goal has been completed. Let
$\mathcal{X}_i(g_{i,k})$ denote the valid navigable
observation positions associated with all instances of the
current target object category. We compute the stage-wise
reference distance as
\begin{equation}
    d^{*}_{i,k}
    =
    \min_{\mathbf{x}\in\mathcal{X}_i(g_{i,k})}
    d_{\mathrm{geo}}
    \left(
        \mathbf{x}^{\mathrm{start}}_{i,k},
        \mathbf{x}
    \right),
    \label{eq:posthoc_distance}
\end{equation}
where $d_{\mathrm{geo}}$ denotes geodesic distance on the
navigation mesh. This reference depends only on the actual
stage start, the current object goal, and its valid target
instances; it does not use future object goals. Different
agents can therefore be evaluated using shortest paths
consistent with the target object instances selected during
their preceding subtasks.

\noindent\textbf{Per-Goal and Sequence-Level Metrics.}
Let $\mathrm{SR}_{i,k}\in\{0,1\}$ indicate whether episode
$i$ successfully completes its $k$-th object-goal subtask,
and let $l_{i,k}$ denote the path length executed during that
subtask. Its per-goal path efficiency is
\begin{equation}
    \mathrm{SPL}_{i,k}
    =
    \mathrm{SR}_{i,k}
    \frac{
        d^{*}_{i,k}
    }{
        \max\left(d^{*}_{i,k},l_{i,k}\right)
    }.
    \label{eq:per_goal_spl}
\end{equation}
For stage-wise dataset aggregation, an unexecuted subtask is
assigned zero SR and SPL, while it remains distinguishable
from an executed failure in the recorded episode results.

Following MultiON \cite{wani2020multion}, we report
sequence-level Success Rate (SR), Success weighted by Path
Length (SPL), Progress Rate (PR), and Progress weighted by
Path Length (PPL). PR measures the fraction of completed object goals, while
PPL weights the successfully completed prefix by its path
efficiency.

\noindent\textbf{Stage-Wise Conditional Evaluation.}
Sequence-level metrics do not reveal how performance changes
across successive object goals. Inspired by the stage-wise
analysis of semantic-memory reuse in OneMap
\cite{busch2025onemap}, we therefore report Conditional SR
and Conditional SPL for each object-goal index.

Under our sequential termination protocol, the $k$-th
subtask is executed only when all preceding object goals have
been completed. By defining
$\mathrm{SR}_{i,0}=1$, the number of episodes that reach
stage $k$ is $\sum_{i=1}^{N}\mathrm{SR}_{i,k-1}$.
Conditional performance is therefore
\begin{equation}
    \mathrm{SR}^{\mathrm{cond}}_{k}
    =
    \frac{
        \sum_{i=1}^{N}\mathrm{SR}_{i,k}
    }{
        \sum_{i=1}^{N}\mathrm{SR}_{i,k-1}
    },
    \qquad
    \mathrm{SR}_{i,0}=1,
    \label{eq:conditional_sr}
\end{equation}
and
\begin{equation}
    \mathrm{SPL}^{\mathrm{cond}}_{k}
    =
    \frac{
        \sum_{i=1}^{N}\mathrm{SPL}_{i,k}
    }{
        \sum_{i=1}^{N}\mathrm{SR}_{i,k-1}
    }.
    \label{eq:conditional_spl}
\end{equation}
Conditional metrics evaluate success and efficiency only among
episodes that reach stage $k$.
Conditional SPL is particularly useful for examining
whether later object goals are reached more efficiently as
geometric and semantic memory accumulates.

For completeness, the stage-wise Global metrics average
over all $N$ episodes, with unexecuted subtasks contributing
zero:
\begin{equation}
    \mathrm{SR}^{\mathrm{global}}_k
    =
    \frac{1}{N}
    \sum_{i=1}^{N}\mathrm{SR}_{i,k},
    \qquad
    \mathrm{SPL}^{\mathrm{global}}_k
    =
    \frac{1}{N}
    \sum_{i=1}^{N}\mathrm{SPL}_{i,k}.
    \label{eq:global_metrics}
\end{equation}
Under the sequential termination protocol, these metrics are
directly related to their Conditional counterparts:
\begin{equation}
\begin{aligned}
    \mathrm{SR}^{\mathrm{global}}_k
    &=
    \mathrm{SR}^{\mathrm{global}}_{k-1}
    \mathrm{SR}^{\mathrm{cond}}_k,\\
    \mathrm{SPL}^{\mathrm{global}}_k
    &=
    \mathrm{SR}^{\mathrm{global}}_{k-1}
    \mathrm{SPL}^{\mathrm{cond}}_k,
\end{aligned}
\qquad
\mathrm{SR}^{\mathrm{global}}_0=1.
\label{eq:conditional_global_relation}
\end{equation}

Global metrics additionally incorporate the probability of
reaching stage $k$ and are used only to distinguish
stage-conditioned performance from preceding failures.
We focus on Conditional SR and Conditional SPL in the main
text and report Global results in the supplementary material.
Sequence-level SR, SPL, PR, and PPL summarize complete and
partial episode progress.

\subsection{Baselines and Evaluation Settings}
\noindent\textbf{Multi-Object Navigation.}
We adopt OneMap~\cite{busch2025onemap} as the primary
baseline for sequential multi-object navigation, as it
maintains a persistent open-vocabulary 2D semantic map
across successive object goals. However, OneMap does not
explicitly support cross-floor navigation. Existing
cross-floor ObjectNav methods primarily focus on
single-object tasks, leaving no directly comparable baseline
that jointly supports persistent multi-object navigation
and explicit floor transitions.

To evaluate the contribution of cross-target memory,
we introduce \textbf{w/o CT Memory}, which resets the
geometric and semantic maps between successive object
goals. We further evaluate three perception configurations
of LifelongCrossNav: YOLOv7, D-FINE-X, and D-FINE-X
with a final semantic verification gate while retaining the other perception and
navigation components. All methods are evaluated on
the same HM3D-MFMON episodes using the post-hoc
evaluation metrics described above.

\noindent\textbf{Single-Object Navigation.}
As an auxiliary evaluation, we follow the OneMap setting
and evaluate LifelongCrossNav on the HM3D ObjectNav
validation split, which contains 2,000 episodes over six
target object categories~\cite{yadav2023hm3dsem}. We
compare against representative task-specific, zero-shot,
open-vocabulary, and floor-aware ObjectNav methods,
including SGM~\cite{zhang2024sgm}, XGX~\cite{wasserman2024xgx}, ZSON~\cite{majumdar2022zson}, VLFM~\cite{yokoyama2024vlfm}, SG-Nav~\cite{yin2024sgnav}, OpenFMNav~\cite{kuang2024openfmnav},
OneMap~\cite{busch2025onemap}, InstructNav~\cite{long2025instructnav}, ApexNav~\cite{zhang2025apexnav}, BeliefMapNav~\cite{zhou2025beliefmapnav}, MFNP~\cite{zhang2025mfnp}, and
ASCENT~\cite{gong2026ascent}.
Each episode contains one object goal, and performance is
measured using standard SR and SPL. The experiments follow
the six-category HM3D ObjectNav setting.
All evaluations are conducted on a single NVIDIA RTX 5090
GPU.

\section{Experimental Results}

\subsection{Multi-Object Navigation}

\noindent\textbf{Overall Performance.}
As shown in Table~\ref{tab:all_overall}, LifelongCrossNav
outperforms OneMap across all four metrics, demonstrating
its effectiveness in sequential multi-object navigation
in multi-floor environments.
Compared with w/o CT Memory, the full method achieves
higher SPL and PPL, along with modest improvements in
SR and PR. 
These results suggest that persistent cross-target memory
not only improves navigation efficiency but may also
reduce failures associated with redundant exploration.

\begin{table}[t]
    \centering
    \scriptsize
    \setlength{\tabcolsep}{3.2pt}
    \caption{
    Overall multi-object navigation results on all 927
    HM3D-MFMON episodes.
    }
    \label{tab:all_overall}
    \resizebox{0.8\columnwidth}{!}{
    \begin{tabular}{lcccc}
        \toprule
        Method
        & SR $\uparrow$
        & SPL $\uparrow$
        & PR $\uparrow$
        & PPL $\uparrow$ \\
        \midrule
        OneMap
        & 16.83 & 6.79 & 33.69 & 13.95 \\
        w/o CT Memory
        & 36.03 & 11.46
        & 53.87 & 18.00 \\
        LifelongCrossNav
        & \textbf{36.57} & \textbf{12.38}
        & \textbf{54.33} & \textbf{18.86} \\
        \bottomrule
    \end{tabular}
    }
   
\end{table}

\begin{figure}[t]
    \centering

    \includegraphics[
        width=0.9\columnwidth
    ]{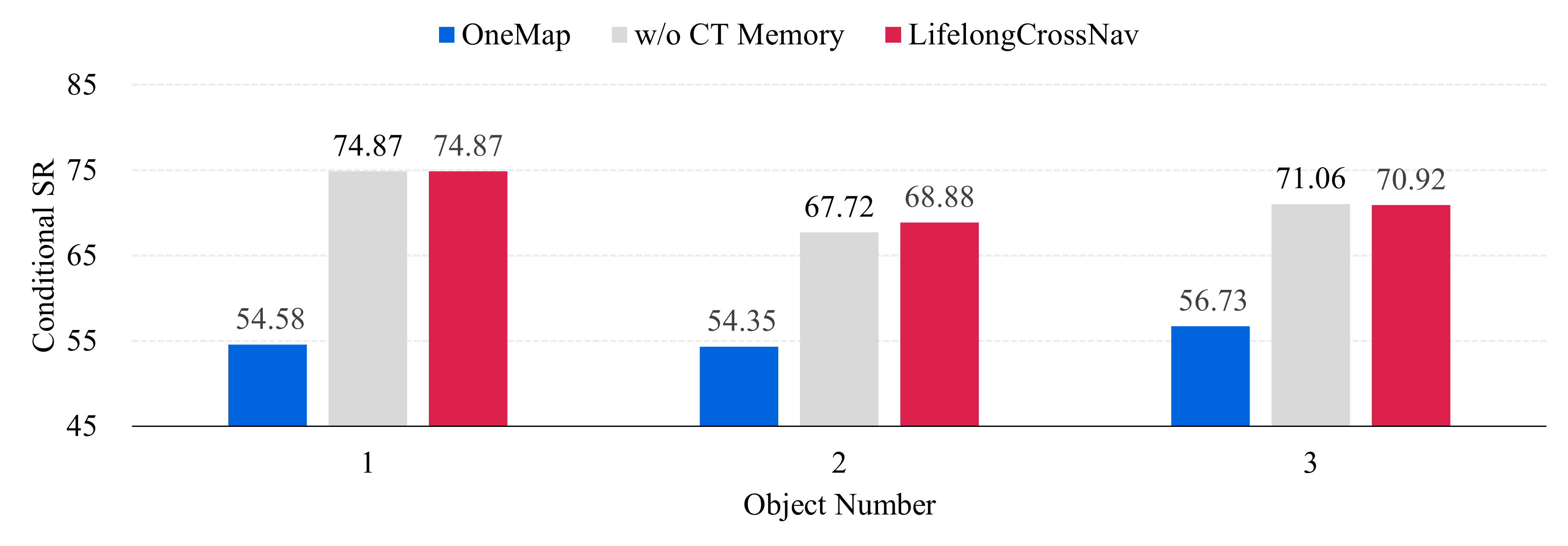}

    \vspace{2pt}

    \includegraphics[
        width=0.9\columnwidth
    ]{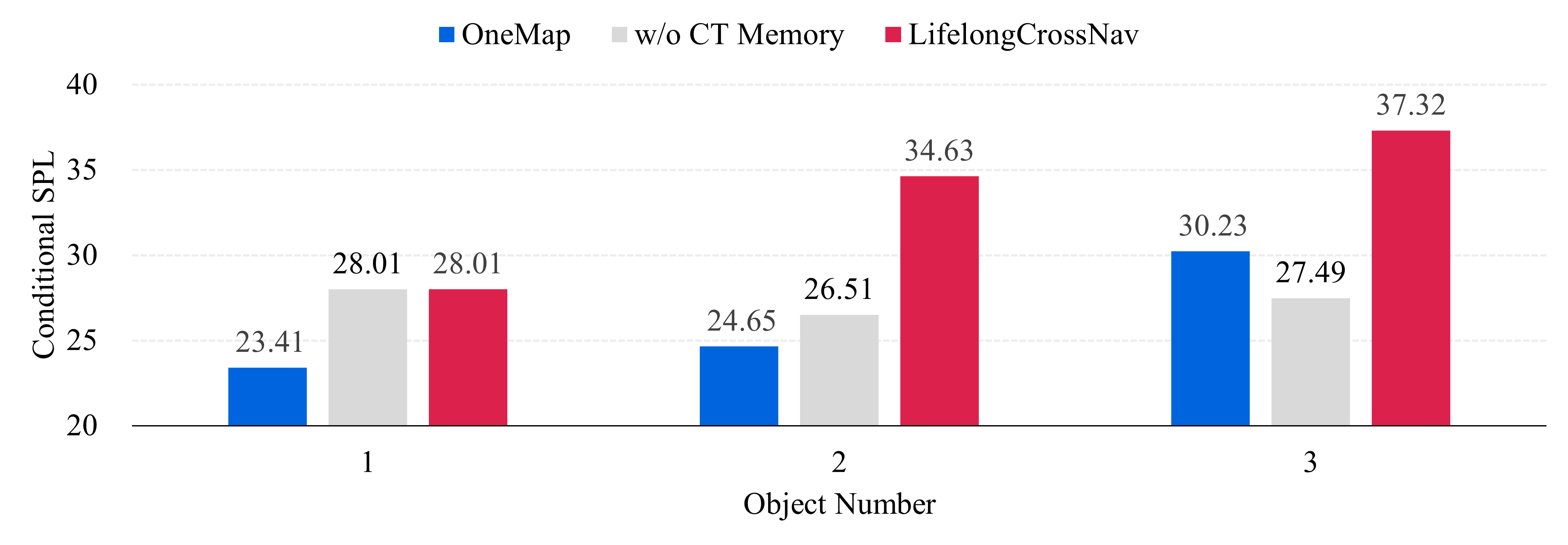}

    \caption{
        Stage-wise conditional performance on all 927
        HM3D-MFMON episodes.
    }
    \label{fig:all_conditional}
    \Description{
        Two bar charts showing Conditional SR (top)
        and Conditional SPL (bottom) across three
        sequential object goals. The charts compare
        OneMap, LifelongCrossNav without cross-target
        memory, and the complete LifelongCrossNav.
    }
\end{figure}

\noindent\textbf{Stage-Wise Results on All Episodes.}
Figure~\ref{fig:all_conditional} reports Conditional SR and
Conditional SPL for the three sequential object goals.
LifelongCrossNav maintains higher Conditional SR than
OneMap throughout the sequence, demonstrating stronger
goal-completion performance in multi-floor environments.
Both persistent-memory methods exhibit higher Conditional
SPL on later goals, whereas the w/o CT Memory variant
shows little improvement.
Notably, the full method and w/o CT Memory perform
identically on the first goal, but diverge on subsequent
goals. This contrast suggests that retaining geometric
and semantic scene memory facilitates historical information
reuse and reduces redundant exploration, improving
later-stage navigation efficiency.

\noindent\textbf{Perception and Verification Ablation.}
As shown in Table~\ref{tab:perception_ablation},
replacing YOLOv7 with D-FINE-X improves all four navigation
metrics, while the final semantic verification gate provides
further gains, particularly in sequence completion and
partial progress. Category-wise analysis also shows fewer
false target confirmations, especially for \texttt{bed}.
Unlike single-object navigation, failures in sequential
multi-object navigation can prevent all subsequent subtasks
from being executed. Thus, improvements in target perception
and verification can mitigate cascading failures, yielding
benefits beyond individual object-goal subtasks.

\begin{table}[t]
    \centering
    \scriptsize
    \setlength{\tabcolsep}{3.2pt}
    \renewcommand{\arraystretch}{1.08}
    \caption{
        Perception and verification ablation on all 927
        HM3D-MFMON episodes.
    }
    \label{tab:perception_ablation}
    \resizebox{\columnwidth}{!}{
    \begin{tabular}{llcccc}
        \toprule
        \multirow{2}{*}{Method}
        & \multirow{2}{*}{Perception Configuration}
        & \multicolumn{4}{c}{HM3D-MFMON} \\
        \cmidrule(lr){3-6}
        & & SR $\uparrow$
          & SPL $\uparrow$
          & PR $\uparrow$
          & PPL $\uparrow$ \\
        \midrule

        \multirow{3}{*}{LifelongCrossNav}
        & YOLOv7
        & 29.13 & 9.64 & 48.40 & 16.70 \\

        & D-FINE-X
        & 31.82 & 11.27 & 50.59 & 18.22 \\

        & D-FINE-X + Sem Gate
        & \textbf{36.57} & \textbf{12.38}
        & \textbf{54.33} & \textbf{18.86} \\

        \bottomrule
    \end{tabular}
    }
\end{table}

\noindent\textbf{Cross-Floor-Required Evaluation.} As shown in Table~\ref{tab:cfr_overall}, OneMap can
complete some preceding subtasks but cannot complete any
full Cross-Floor-Required sequence. LifelongCrossNav
achieves nonzero sequence success and stronger partial
progress, demonstrating that persistent planar memory alone
is insufficient when task completion requires explicit stair
traversal and cross-floor connectivity.

\begin{figure}[t]
    \centering

    \includegraphics[
        width=0.9\columnwidth
    ]{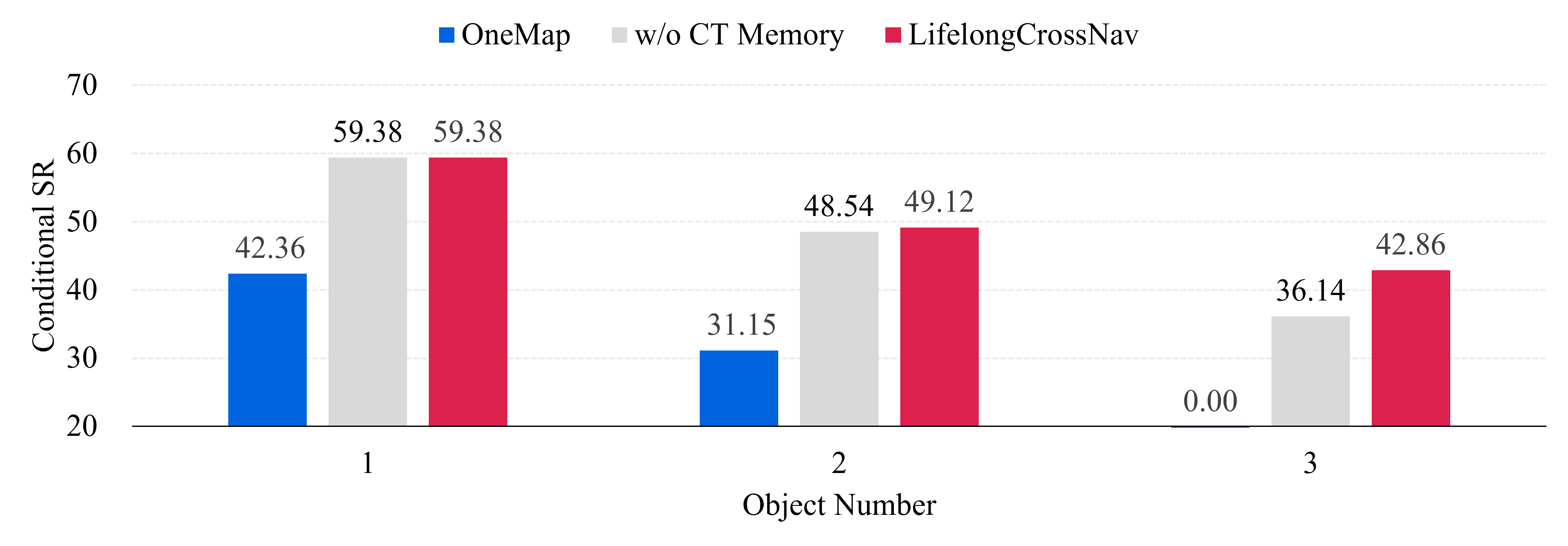}

    \vspace{2pt}

    \includegraphics[
        width=0.9\columnwidth
    ]{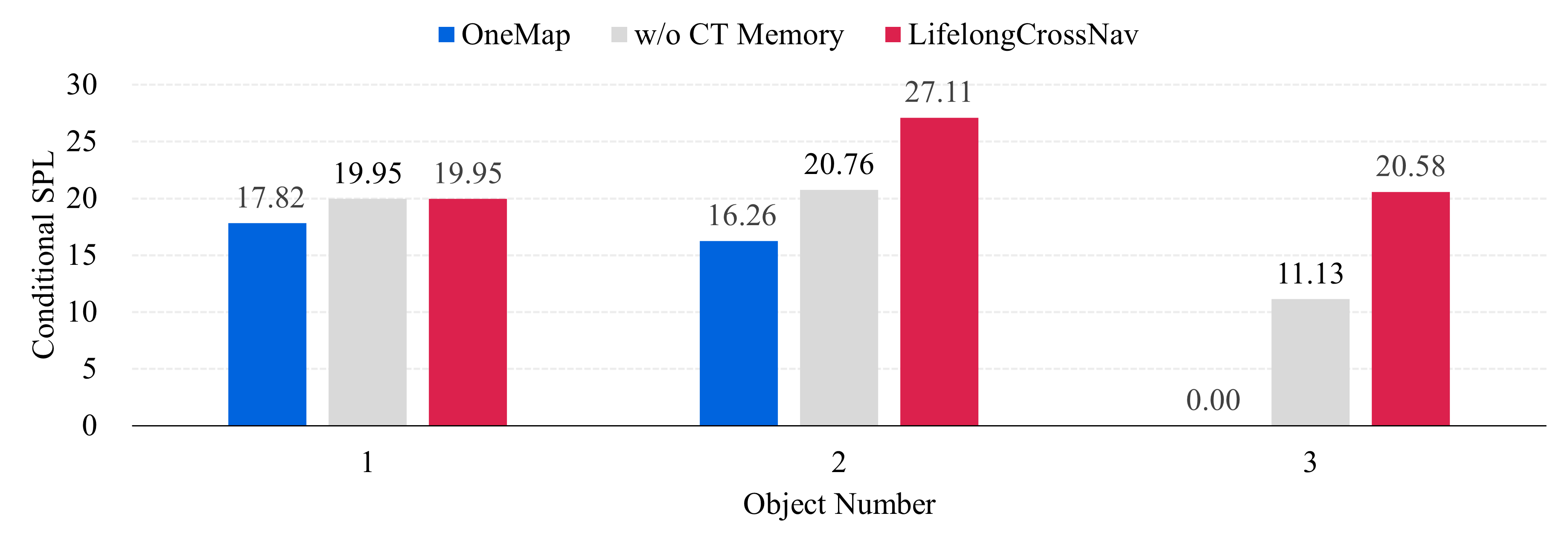}

    \caption{
        Stage-wise conditional performance on the 288
        Cross-Floor-Required episodes.
    }
    \label{fig:cfr_conditional}
    \Description{
        Two bar charts showing Conditional SR (top)
        and Conditional SPL (bottom) across three
        sequential object goals in the CFR subset.
        The charts compare OneMap, LifelongCrossNav
        without cross-target memory, and the complete
        LifelongCrossNav.
    }
\end{figure}

The stage-wise results in Fig.~\ref{fig:cfr_conditional}
show that OneMap cannot complete the remaining object-goal
subtasks once cross-floor traversal becomes necessary. In the
CFR subset, episodes that successfully reach later stages have
already completed previous same-floor subtasks; therefore, the
remaining target is more likely to require the unresolved floor
transition. Without explicit stair representation and
cross-floor connectivity, the planar semantic map cannot
provide an executable solution for these cases.

Unlike the trend over all episodes, Conditional SPL does not
increase monotonically on the CFR subset, as later goals
may still require unseen-floor exploration and additional
stair traversal. Nevertheless, the complete framework remains
more efficient than both OneMap and w/o CT Memory on later
stages, indicating that retaining geometric and semantic
scene memory improves later-stage navigation efficiency
even when floor transitions are required.

\begin{table}[t]
    \centering
    \scriptsize
    \setlength{\tabcolsep}{3.2pt}
    \caption{
    Overall multi-object navigation results on the 288
    Cross-Floor-Required episodes.
    }
    \label{tab:cfr_overall}
    \resizebox{0.8\columnwidth}{!}{
    \begin{tabular}{lcccc}
        \toprule
        Method
        & SR $\uparrow$
        & SPL $\uparrow$
        & PR $\uparrow$
        & PPL $\uparrow$ \\
        \midrule
        OneMap
        & 0.00 & 0.00 & 18.52 & 7.97 \\
        w/o CT Memory
        & 10.42 & 2.23 & 32.87 & 10.21 \\
        LifelongCrossNav
        & \textbf{12.50} & \textbf{3.11}
        & \textbf{33.68} & \textbf{10.86} \\
        \bottomrule
    \end{tabular}
    }

\end{table}

\subsection{Single-Object Navigation}

\begin{table}[t]
    \centering
    \scriptsize
    \setlength{\tabcolsep}{2.3pt}
    \caption{
    Single-object navigation results on HM3D.
    }
    \label{tab:single_object_results}
    \resizebox{\columnwidth}{!}{
    \begin{tabular}{lccccc}
        \toprule
        Method
        & Zero-shot
        & Multi-Floor
        & LLM Reasoning
        & SR $\uparrow$
        & SPL $\uparrow$ \\
        \midrule
        SGM
        & $\times$ & $\times$ & $\times$
        & 60.2 & 30.8 \\
        XGX
        & $\times$ & $\times$ & $\times$
        & \textbf{72.9} & \textbf{35.7} \\
        \midrule
        ZSON
        & $\checkmark$ & $\times$ & $\times$
        & 25.5 & 12.6 \\
        VLFM
        & $\checkmark$ & $\times$ & $\times$
        & 52.5 & 30.4 \\
        SG-Nav
        & $\checkmark$ & $\times$ & $\checkmark$
        & 54.0 & 24.9 \\
        OpenFMNav
        & $\checkmark$ & $\times$ & $\checkmark$
        & 54.9 & 24.4 \\
        OneMap
        & $\checkmark$ & $\times$ & $\times$
        & 55.8 & \textbf{37.4} \\
        InstructNav
        & $\checkmark$ & $\times$ & $\checkmark$
        & 58.0 & 20.9 \\
        ApexNav
        & $\checkmark$ & $\times$ & $\checkmark$
        & 59.6 & 33.0 \\
        BeliefMapNav
        & $\checkmark$ & $\times$ & $\checkmark$
        & \textbf{61.4} & 30.6 \\
        \midrule
        MFNP
        & $\checkmark$ & $\checkmark$ & $\checkmark$
        & 58.3 & 26.7 \\
        ASCENT
        & $\checkmark$ & $\checkmark$ & $\checkmark$
        & \textbf{65.4} & \textbf{33.5} \\
        \midrule
        \textbf{LifelongCrossNav}
        & $\checkmark$ & $\checkmark$ & $\times$
        & \textbf{59.9} & \textbf{26.9} \\
        \bottomrule
    \end{tabular}
    }
    
\end{table}




Table~\ref{tab:single_object_results} shows that
LifelongCrossNav achieves competitive single-object
navigation performance without online LLM reasoning.
OneMap also operates without online LLM reasoning.
Compared with OneMap, LifelongCrossNav achieves higher
SR but lower SPL. This may partly result from its
current-floor-first strategy, which prioritizes local
exploration before searching other floors.


\subsection{Failure Analysis}

\begin{figure}[t]
    \centering

    \begin{minipage}[t]{0.49\linewidth}
        \centering
        \includegraphics[width=\linewidth]{
            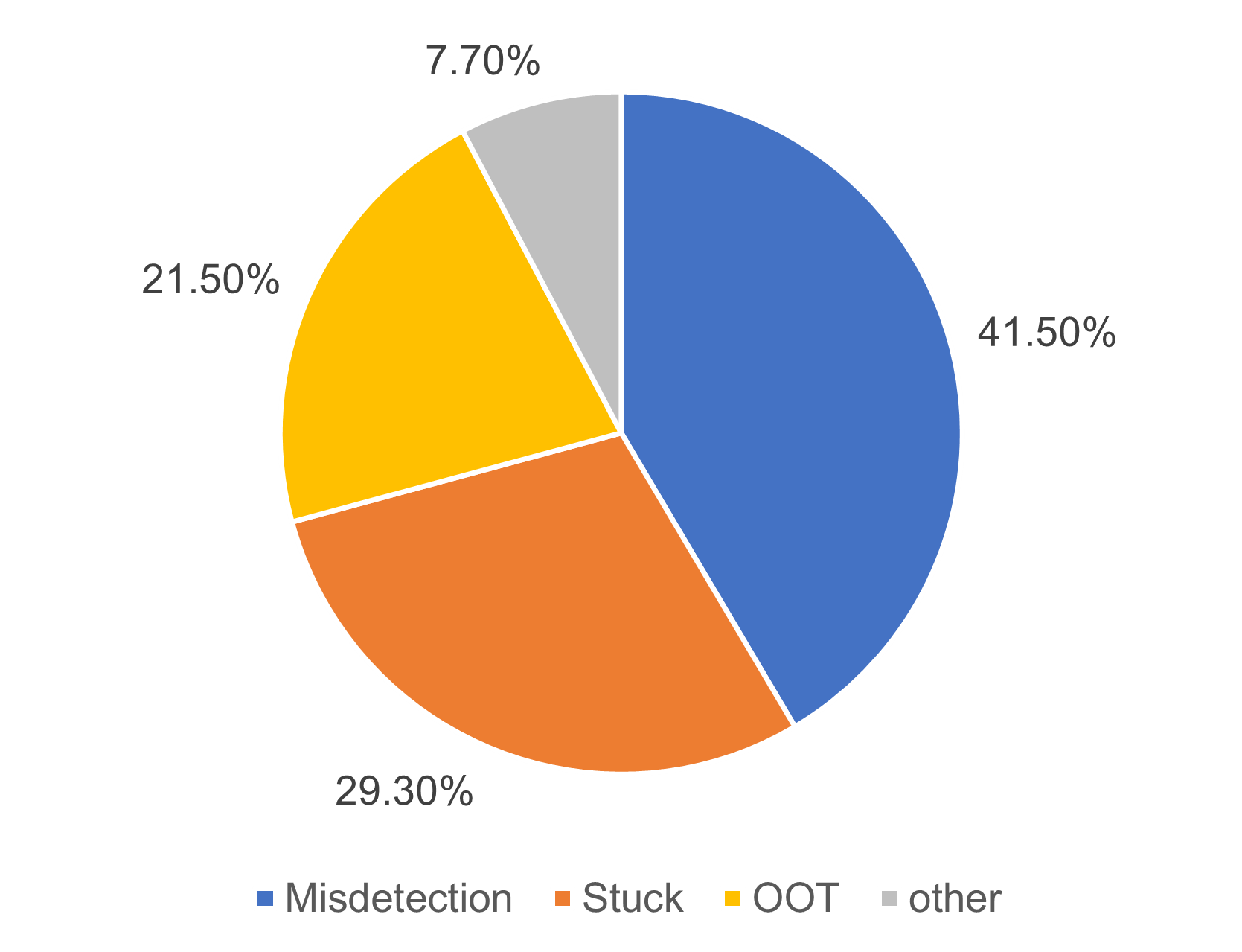}
        \par{\scriptsize (a) Single-object}
    \end{minipage}%
    \hfill
    \begin{minipage}[t]{0.49\linewidth}
        \centering
        \includegraphics[width=\linewidth]{
            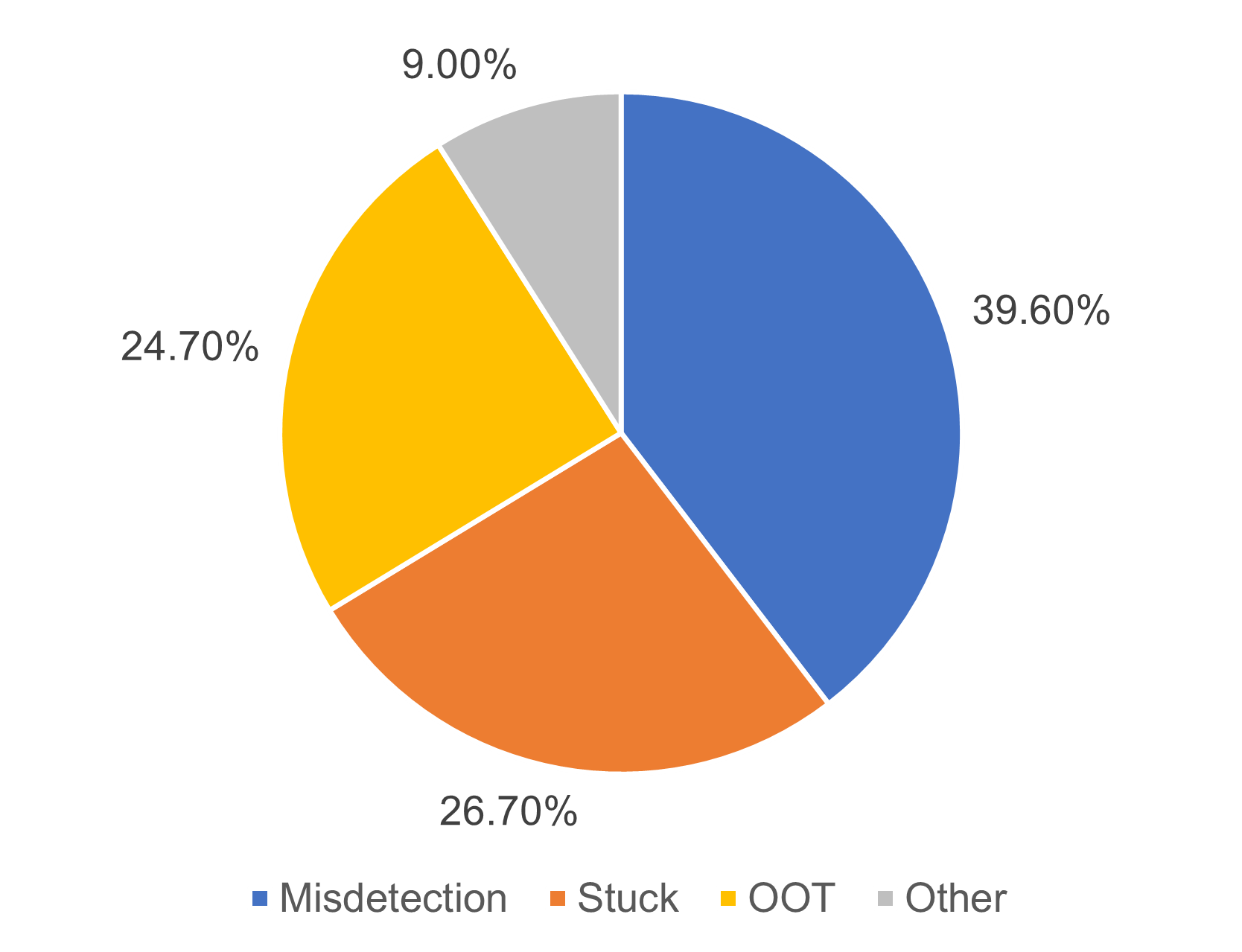}
        \par{\scriptsize (b) Multi-object}
    \end{minipage}

    \par\vspace{4pt}

    \begin{minipage}[t]{0.49\linewidth}
        \centering
        \includegraphics[width=\linewidth]{
            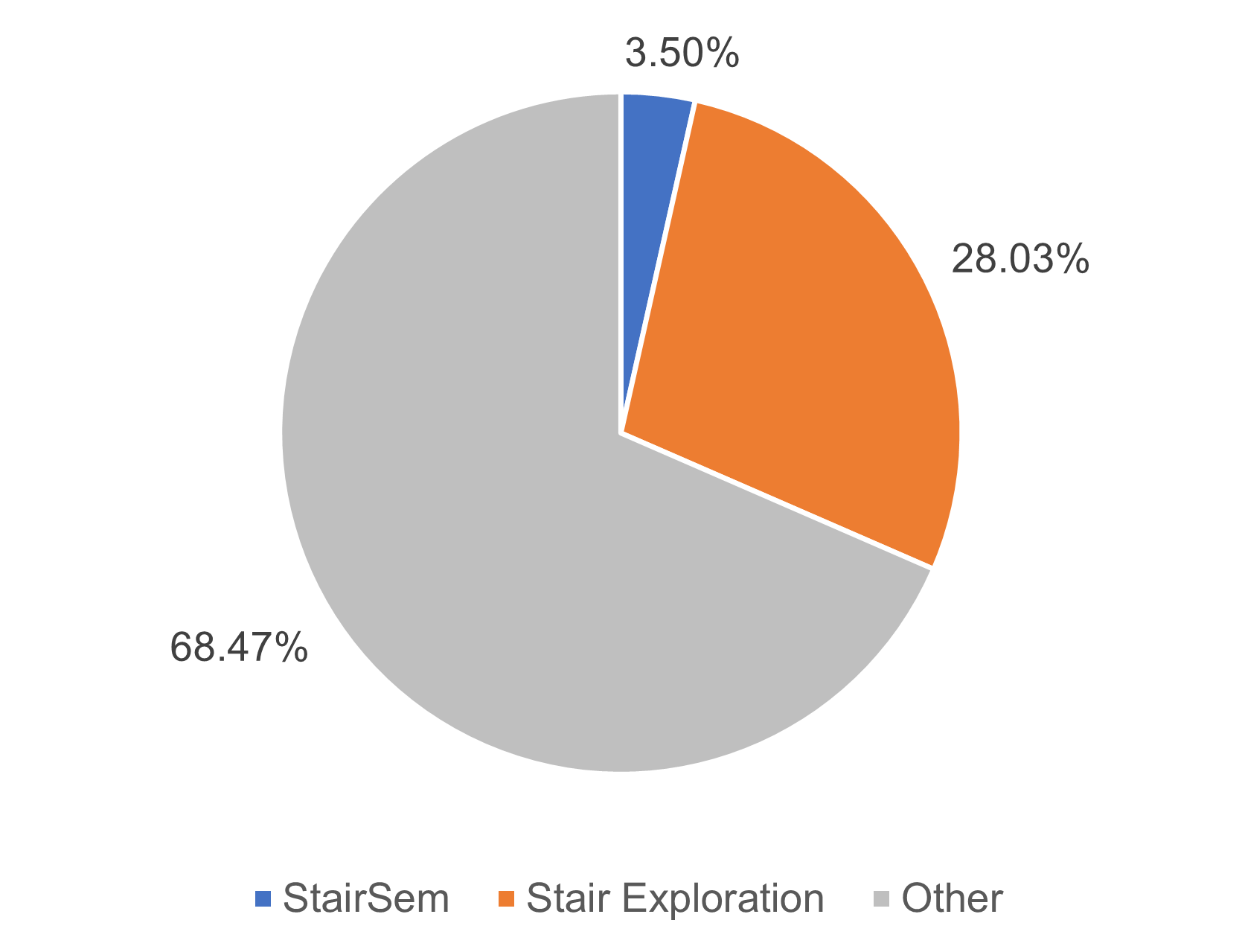}
        \par{\scriptsize (c) Single-object cross-floor}
    \end{minipage}%
    \hfill
    \begin{minipage}[t]{0.49\linewidth}
        \centering
        \includegraphics[width=\linewidth]{
            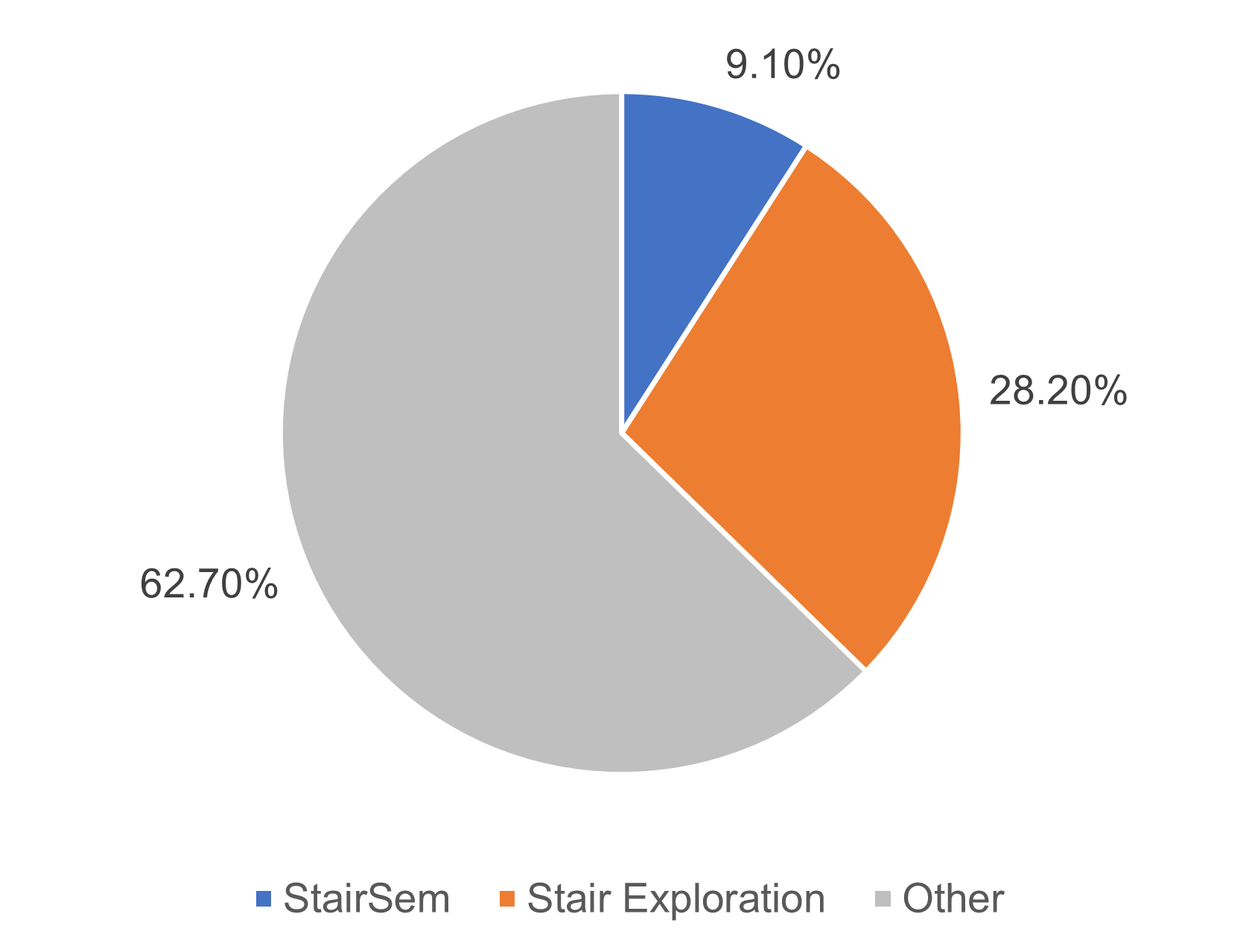}
        \par{\scriptsize (d) Multi-object CFR}
    \end{minipage}

    \caption{
        Failure analysis of LifelongCrossNav.
        (a--b) Overall failure distributions in single-object
        and multi-object navigation.
        (c--d) Terminal navigation modes in failed
        cross-floor tasks.
    }
    \label{fig:failure_analysis}
    \Description{
        Four pie charts arranged in a two-by-two grid.
        The upper charts show the proportions of
        misdetection, navigation stalls, timeouts, and
        other failures in single-object and multi-object
        navigation. The lower charts show terminal
        navigation modes for failed single-object
        cross-floor episodes and multi-object CFR episodes,
        grouped into StairSem, stair-directed exploration,
        and other modes.
    }
\end{figure}

\noindent\textbf{Perception Failures.}
As shown in Fig.~\ref{fig:failure_analysis}(a--b),
target misdetection is the most frequent failure type,
accounting for 41.5\% and 39.6\% of failures in
single-object and multi-object navigation, respectively.
These errors are particularly common for \texttt{bed}
and \texttt{sofa}. In sequential multi-object navigation,
a misdetection can terminate the current subtask and
prevent subsequent goals from being executed.
Such cascading failures amplify the impact of perception
errors on sequence completion, highlighting target
verification as an important direction for future improvement.

\noindent\textbf{Stair-Related Navigation Failures.}
Figure~\ref{fig:failure_analysis}(c--d) shows the terminal
navigation modes in cross-floor tasks.
In failed single-object cross-floor episodes, 28.0\%
terminate during stair-directed exploration, compared
with only 3.5\% in StairSem mode.
A similar pattern appears in multi-object CFR episodes,
with 28.2\% and 9.1\%, respectively.
Failures during stair-directed exploration are mainly
associated with navigation stalls and timeouts.
These results suggest that difficulties often arise
during stair discovery and approach, before entering
StairSem mode. This may be related to incomplete stair
perception and limitations in mesh geometry and
traversability representation near stair entrances.

\section{Conclusion}

We presented LifelongCrossNav for sequential multi-object
ObjectNav in multi-floor environments, integrating persistent
3D semantic memory, support-aware traversability, and
stair-aware navigation. We also introduced HM3D-MFMON and
a post-hoc protocol for stage-wise evaluation. Experiments
demonstrate cross-floor sequence completion beyond a planar
semantic-memory baseline, while cross-target memory improves
later-goal path efficiency and SED verification enhances
target confirmation. Future work will focus on more robust stair exploration, adaptive floor selection, and semantic target verification.
\bibliographystyle{ACM-Reference-Format}
\bibliography{references_lifelongcrossnav}

\clearpage
\appendix
\section*{Supplementary Material}
This supplement provides the configuration, voxel-state definitions,
navigation-mode details, type-aware planning rules, and additional stage-wise
evaluation results omitted from the main paper.

\section{Computational Environment}
\label{sec:supp_environment}

All reported experiments were conducted using the software and hardware
environment summarized in Table~\ref{tab:supp_environment}. The CUDA version
listed below refers to the runtime bundled with PyTorch, rather than the
version reported by the host GPU driver.

\section{Experimental Configuration}
\label{sec:supp_configuration}

Each HM3D-MFMON episode contains three sequential object-goal queries. The
sparse geometric map and goal-independent SED/CLIP features persist across
queries within the same episode. Query-conditioned similarity scores, POIs,
active paths, frontier selections, and controller states are refreshed when
the active object goal changes. All states are reset between independent
episodes.

Table~\ref{tab:supp_configuration} summarizes the frozen settings used for the
reported LifelongCrossNav experiments. The 1500-action budget is shared by the
three object-goal subtasks within each episode. The internal target-approach
range is a policy parameter, whereas the evaluator success radius is used only
for ground-truth success assessment. The full model uses D-FINE-X for target
detection and a final SED verification threshold of 0.15.

\section{Persistent 3D Semantic Memory}
\label{sec:supp_memory}

\subsection{Dense Feature Projection and Fusion}

For each RGB frame, the SED image encoder produces a spatial feature tensor
\begin{equation}
\mathbf F_t^{\mathrm{dense}}
\in\mathbb R^{24\times24\times768}.
\label{eq:supp_dense_feature}
\end{equation}
The feature tensor is bilinearly interpolated to the depth resolution. Each
valid depth pixel is then back-projected with the camera intrinsics and
transformed into the global frame using the 6-DoF camera pose. Consequently,
features are written only to observed surface voxels rather than being copied
uniformly over free space.

For pixel $(u,v)$, the observation quality is
\begin{equation}
\begin{aligned}
q_t(u,v)
={}&
\left[
1-\tanh\!\left(
\gamma\left\|\nabla D_t(u,v)\right\|
\right)
\right] \\
&\times
\exp\!\left[
-\frac{(d_0-D_t(u,v))^2}{3\sigma_d^2}
\right].
\end{aligned}
\label{eq:supp_observation_quality}
\end{equation}
Depth discontinuities and observations far from the preferred range therefore
receive smaller weights.

If voxel $v$ has accumulated feature $\mathbf f_v^{t-1}$ and confidence
$w_v^{t-1}$, the new observation is fused using
\begin{equation}
\begin{aligned}
\mathbf f_v^t
&=
\frac{
w_v^{t-1}\mathbf f_v^{t-1}
+
q_v^t\mathbf f_{v,\mathrm{obs}}^t
}{
w_v^{t-1}+q_v^t
},\\
w_v^t
&=
w_v^{t-1}+q_v^t.
\end{aligned}
\label{eq:supp_feature_fusion}
\end{equation}
This is a cumulative weighted average rather than an exponential moving
average. The fused features and confidence values persist across object-goal
subtasks within an episode.

\subsection{Query-Conditioned Retrieval}

For the active object-goal text $q$, the normalized text embedding
$\mathbf t_q$ is compared with every feature-bearing voxel:
\begin{equation}
s_q(v)
=
\frac{
\mathbf f_v^\top\mathbf t_q
}{
\|\mathbf f_v\|_2\|\mathbf t_q\|_2
}.
\label{eq:supp_similarity}
\end{equation}
The resulting 3D similarity field is query-dependent and is not part of the
persistent memory itself. Spatially consistent high-similarity responses are
clustered, and the four highest-ranked clusters form the initial History-POI
candidate set.

When the active object goal changes, the stored voxel features remain
unchanged, whereas the similarity field and History POIs are recomputed for
the new text query.

\begin{table}[t]
\centering
\caption{Software and hardware environment used for the reported experiments.}
\label{tab:supp_environment}
\scriptsize
\setlength{\tabcolsep}{4.0pt}
\renewcommand{\arraystretch}{1.08}
\begin{tabular}{p{0.43\columnwidth}p{0.49\columnwidth}}
\toprule
Component & Tested value \\
\midrule
OS
& Ubuntu 22.04, Linux 6.8 \\

Python
& 3.10.20 \\

PyTorch / torchvision
& 2.11.0+cu128 / 0.26.0+cu128 \\

CUDA runtime used by PyTorch
& 12.8 \\

GPU used for paper runs
& NVIDIA GeForce RTX 5090, 32 GB \\

Habitat-Sim / Habitat-Lab
& 0.2.4 / 0.2.4 \\

Transformers
& 4.26.1 \\

OpenCV / NumPy
& 4.8.0 / 1.26.4 \\

Detectron2 / timm
& 0.6 / 1.0.26 \\

Rerun SDK
& 0.23.1 \\
\bottomrule
\end{tabular}
\end{table}

\begin{table*}[t]
\centering
\caption{Frozen configuration used for the reported
LifelongCrossNav experiments.}
\label{tab:supp_configuration}
\scriptsize
\setlength{\tabcolsep}{4.0pt}
\renewcommand{\arraystretch}{1.08}
\begin{tabular}{llr@{\hspace{7pt}}llr}
\toprule
Group & Parameter & Value
& Group & Parameter & Value \\
\midrule
Sensor
& RGB/depth resolution
& $640\!\times\!640$
& Map
& Metric XY extent
& $60\!\times\!60$ m \\

Sensor
& Horizontal field of view
& $90^\circ$
& Map
& Voxel resolution (XYZ)
& 0.10 m \\

Sensor
& Camera height
& 0.88 m
& Map
& Indexed Z extent
& approx. $\pm30$ m \\

Action
& Forward step
& 0.25 m
& Map
& Inflation radius
& 0.20 m \\

Action
& Yaw/pitch increment
& $30^\circ$
& Map
& Support search depth
& 0.40 m \\
\midrule

SED
& Encoder input
& $768\!\times\!768$
& SED
& Dense output
& $24\!\times\!24\!\times\!768$ \\

Fusion
& Depth-gradient factor $\gamma$
& 0.20
& Fusion
& Preferred depth $d_0$
& 1.0 m \\

Fusion
& Depth scale $\sigma_d$
& 1.5 m
& History POI
& Retrieved clusters
& Top-4 \\

History POI
& Blacklist cylinder
& 0.8 m XY, 0.6 m Z
& POI
& Live takeover radius
& 1.0 m \\
\midrule

Object
& Detector (full model)
& D-FINE-X
& Object
& Secondary confidence
& 0.60 \\

Object
& Found confidence
& 0.70
& Object
& Minimum box area
& $1/64$ image \\

Object
& Minimum projected voxels
& 10
& Object
& Target-approach range
& 1.30 m \\

Verification
& Final SED threshold
& 0.15
& Verification
& Mask similarity pooling
& Top 20\% \\

Stair
& SegFormer pixel threshold
& 0.25
& Stair
& Minimum component area
& 80 px \\

Stair
& Descent confirmation
& 0.48
& Stair
& Geometry threshold
& 0.65 \\
\midrule

Frontier
& Normal minimum cluster
& 5
& Frontier
& Init./descent minimum cluster
& 2 \\

StairSem
& Session admission radius
& 2.5 m
& StairSem
& Arrival threshold
& 0.40 m \\

Planner
& Normal graph
& 26-neighbor
& Planner
& StairSem edge radius
& 1.0 m \\
\midrule

Evaluation
& Episode action budget
& 1500
& Evaluation
& Success radius
& 1.50 m \\

Navmesh
& Agent height/radius
& 1.50/0.10 m
& Navmesh
& Maximum climb/slope
& 0.30 m/$60^\circ$ \\

Navmesh
& Cell size/height
& 0.03/0.05 m
& Dataset
& Goals/categories
& 3/6 \\
\bottomrule
\end{tabular}
\end{table*}

\section{Support-Aware Voxel States}
\label{sec:supp_voxel_states}

The sparse map maintains seven operational voxel states. Surface states store
observed geometry, while air states represent the volume occupied by the
agent above a supporting surface. Table~\ref{tab:supp_voxel_states} summarizes
their construction and planning roles.

\begin{table*}[t]
\centering
\caption{Operational voxel states and their roles in mapping and planning.}
\label{tab:supp_voxel_states}
\scriptsize
\setlength{\tabcolsep}{3.2pt}
\renewcommand{\arraystretch}{1.12}
\begin{tabular}{
l
p{0.45\textwidth}
c
c
p{0.16\textwidth}}
\toprule
Voxel state
& Construction and geometric interpretation
& Pathable
& Targetable
& Planning role \\
\midrule

\texttt{OCCUPIED}
& Surface voxel produced by a valid RGB-D depth return. It represents walls,
furniture, ordinary floors, and other observed geometry and may carry a
semantic feature.
& No
& No
& Collision geometry and ordinary support. \\

\texttt{TRAVERSABLE}
& Ray-observed free-space voxel whose first valid support within 0.40 m is
\texttt{OCCUPIED} or reversible \texttt{FAKE\_STAIR}.
& Yes
& Yes
& Primary state for same-floor navigation. \\

\texttt{TRAVERSABLE\_STAIR}
& Free-space voxel whose first valid support is \texttt{STAIR} or
\texttt{STAIR\_EXPANDED}.
& Yes
& Yes
& Robot occupancy above a confirmed stair surface. \\

\texttt{TRAVERSABLE\_FAKE}
& Ray-observed free-space voxel for which no valid support is found within the
support-search depth.
& No
& No
& Marks unsupported air, drop-offs, or stair voids. \\

\texttt{STAIR}
& Surface voxel accepted as stair geometry after semantic and 3D geometric
verification.
& Yes
& Yes
& Confirmed stair body used during StairSem. \\

\texttt{STAIR\_EXPANDED}
& Neighboring \texttt{OCCUPIED} voxel assimilated into a confirmed stair
component to improve sparse geometric continuity.
& Yes
& Yes
& Complements incomplete stair observations with a higher planning cost. \\

\texttt{FAKE\_STAIR}
& Rejected or reversible stair evidence retained instead of being immediately
discarded.
& Fallback
& No
& High-cost temporary connection and reversible support state. \\
\bottomrule
\end{tabular}
\end{table*}

\subsection{Support Classification}

Depth returns are first inserted as \texttt{OCCUPIED}. Ray casting creates
observed free-space samples, after which support is searched vertically below
each sample. Let $s(v)$ denote the first valid support type found below voxel
$v$. We define the ordinary-support and stair-support sets as
$\mathcal{S}_{\mathrm{ord}}
=\{\texttt{OCCUPIED},\texttt{FAKE\_STAIR}\}$
and
$\mathcal{S}_{\mathrm{stair}}
=\{\texttt{STAIR},\texttt{STAIR\_EXPANDED}\}$,
respectively. The derived air state is then determined by
\begin{equation}
\tau_{\mathrm{air}}(v)=
\begin{cases}
\texttt{TRAVERSABLE},
& s(v)\in\mathcal{S}_{\mathrm{ord}},\\
\texttt{TRAVERSABLE\_STAIR},
& s(v)\in\mathcal{S}_{\mathrm{stair}},\\
\texttt{TRAVERSABLE\_FAKE},
& s(v)=\varnothing.
\end{cases}
\label{eq:supp_support_classification}
\end{equation}

Accordingly, \texttt{TRAVERSABLE} represents supported free space above an
ordinary or reversible surface, whereas \texttt{TRAVERSABLE\_STAIR} denotes
free space supported by confirmed stair geometry. In contrast,
\texttt{TRAVERSABLE\_FAKE} indicates observed free space for which no valid
support is found within the support-search depth. It should therefore not be
confused with \texttt{FAKE\_STAIR}, which is a reversible surface state rather
than an unsupported-air state.

\subsection{Stair-State Promotion}

During Basic Explore, SegFormer masks and RGB-D geometry may generate ascent or
descent evidence, but ordinary \texttt{OCCUPIED} voxels are not immediately
rewritten as confirmed stairs. After the corresponding stair entrance is
reached and verified, the controller enters Stair Explore, implemented as the
StairSem mode.

Within StairSem, accepted surface voxels are promoted to \texttt{STAIR};
adjacent occupied cells may become \texttt{STAIR\_EXPANDED}; rejected evidence
is retained as \texttt{FAKE\_STAIR}. Support-dependent air voxels are then
reclassified locally. This delayed promotion prevents uncertain stair
observations collected during ordinary exploration from directly changing the
cross-floor planning graph.

\section{Candidate Hierarchy and Navigation Modes}
\label{sec:supp_navigation_modes}

\subsection{Candidate Priority}

The controller considers four main candidate classes:
\begin{equation}
\mathcal C_t=
\mathcal P_t^{\mathrm{live/history}}
\cup
\mathcal F_t^{\mathrm{trav}}
\cup
\mathcal F_t^{\mathrm{desc}}
\cup
\mathcal F_t^{\mathrm{asc}}.
\label{eq:supp_candidate_set}
\end{equation}
Their default selection priority is
\begin{equation}
\mathcal P_t^{\mathrm{live/history}}
\succ
\mathcal F_t^{\mathrm{trav}}
\succ
\mathcal F_t^{\mathrm{desc}}
\succ
\mathcal F_t^{\mathrm{asc}}.
\label{eq:supp_candidate_priority}
\end{equation}

Here, Basic Frontiers collectively refer to Traversable, Descend, and Ascent
Frontiers used during ordinary exploration:

\begin{itemize}
    \item \textbf{Traversable Frontier}: the boundary between supported
    \texttt{TRAVERSABLE} voxels and genuine unknown space;
    \item \textbf{Descend Frontier}: the boundary between supported traversable
    space and unsupported or fake-stair geometry;
    \item \textbf{Ascent Frontier}: a semantic-geometric entrance proposal for
    an upward stair.
\end{itemize}

The controller first explores reachable Traversable Frontiers on the current
floor. Only after no reachable ordinary frontier remains does it consider
Descend and Ascent Frontiers. Descend is attempted before ascent by default,
although the order may be adapted according to the most recent successful
floor-transition direction.

A selected candidate is committed for a short grace period so that incremental
map updates do not cause frequent target switching. Commitment is therefore a
stability mechanism rather than an additional candidate class. A committed
candidate is replaced only when it becomes invalid, unreachable, or is
superseded by a higher-priority POI.

\subsection{Mode Definitions}

Table~\ref{tab:supp_navigation_modes} summarizes the high-level modes and
their planning behavior.

\begin{table*}[t]
\centering
\caption{Navigation modes, target types, and mode-dependent planning behavior.}
\label{tab:supp_navigation_modes}
\scriptsize
\setlength{\tabcolsep}{3.0pt}
\renewcommand{\arraystretch}{1.12}
\begin{tabular}{
p{0.15\textwidth}
p{0.20\textwidth}
p{0.18\textwidth}
p{0.27\textwidth}
p{0.14\textwidth}}
\toprule
Mode
& Navigation target
& Graph construction
& Main behavior and admissibility
& Exit condition \\
\midrule

\textbf{Basic Explore}
& Traversable, Descend, or Ascent Frontier
& Strict 26-neighbor graph; selected target remains on the current exploration
level
& Prioritizes ordinary \texttt{TRAVERSABLE} space. Descend and ascent targets
are entrance proposals; Basic Explore does not actively traverse an
unconfirmed stair body.
& Valid POI, verified stair entrance, or exhausted candidates. \\

\textbf{POI Navigation}
& Live POI or History POI
& Standard 26-neighbor 3D graph
& Navigates toward current or previously observed semantic evidence. Previously
confirmed stair connections may be reused to reach a POI on another floor.
& POI arrival, invalidation, live-target takeover, or failure. \\

\textbf{Stair Explore / StairSem}
& Directional Stair Frontier
& Pathable endpoints inside a 1.0 m physical 3D sphere
& Admits confirmed stair states and stair-supported air. The requested ascent
or descent direction constrains vertical progress and suppresses unrelated
platforms or neighboring stair flights.
& Stair-frontier arrival followed by multi-frame landing confirmation. \\

\textbf{Target Approach}
& Observation viewpoint near a Live POI
& Local path planning and discrete view correction
& Refines the camera pose, target mask, and stopping decision. It does not
initiate new cross-floor exploration.
& Valid found action, rejected detection, or return to exploration. \\
\bottomrule
\end{tabular}
\end{table*}

During History-POI navigation, a new Live POI may take control only when its
projected 3D location is spatially consistent with the active History-POI
region. This prevents unrelated detections elsewhere in the scene from
interrupting historical retrieval.

\section{Type-Aware 3D A$^\ast$}
\label{sec:supp_astar}

\subsection{Normal Graph}

Basic Explore and POI Navigation use a 26-neighbor graph. The heuristic is the
3D Euclidean distance from the current voxel to the goal:
\begin{equation}
h(n)
=
\sqrt{
(i_x-g_x)^2+
(i_y-g_y)^2+
(i_z-g_z)^2
}.
\label{eq:supp_astar_heuristic}
\end{equation}

The transition cost is
\begin{equation}
c(n,n')
=
\|n-n'\|_2+p(\tau_{n'}),
\label{eq:supp_astar_edge_cost}
\end{equation}
where $p(\tau_{n'})$ is determined by the destination voxel state.

The penalties favor stable supported air while preserving access to confirmed
stair geometry. \texttt{FAKE\_STAIR} remains available only as a costly
fallback and is never selected as a navigation endpoint.

Ordinary planning prioritizes non-inflated voxels. When a mode explicitly
permits an inflated pathable endpoint, its transition cost is multiplied by
two. If the agent is already inside an inflated pathable region, the planner
may temporarily retain inflated states to allow the agent to leave that
region; strict planning resumes afterward.

\subsection{StairSem Graph}

Sparse RGB-D observations may leave gaps larger than one voxel between
successive stair surfaces. StairSem therefore replaces the fixed 26-neighbor
relation with all existing pathable endpoints inside a 1.0 m physical sphere.
This neighborhood is defined in metric 3D space rather than by a fixed voxel
offset.

Inflated \texttt{TRAVERSABLE} and
\texttt{TRAVERSABLE\_STAIR} endpoints may be used when necessary, whereas
inflated \texttt{STAIR}, \texttt{STAIR\_EXPANDED}, and
\texttt{FAKE\_STAIR} endpoints are rejected. If the final Stair Frontier is
temporarily unreachable, intermediate session-local stair targets are tried in
descending order of vertical progress.

The expanded graph is deliberately permissive for incomplete stair
observations. It verifies the stored endpoints of each edge but does not claim
continuous swept-volume collision certification along every intermediate
sample of a long edge.

\begin{table}[t]
\centering
\caption{Voxel-type penalties used by the 3D A$^\ast$ planner.}
\label{tab:supp_astar_penalties}
\scriptsize
\setlength{\tabcolsep}{5pt}
\begin{tabular}{lc}
\toprule
Destination voxel state & Type penalty \\
\midrule
\texttt{TRAVERSABLE} & 0 \\
\texttt{TRAVERSABLE\_STAIR} & 0 \\
\texttt{STAIR} & 4 \\
\texttt{STAIR\_EXPANDED} & 5 \\
\texttt{FAKE\_STAIR} & 10 \\
\bottomrule
\end{tabular}
\end{table}

\section{Evaluation Protocol and Additional Results}
\label{sec:supp_evaluation}

\subsection{Benchmark Summary}

HM3D-MFMON contains 927 three-goal episodes from 36 multi-floor HM3D scenes.
The finalized Cross-Floor-Required (CFR) subset contains 288 episodes whose
complete object-goal sequences require at least one floor transition. The
first mandatory transition occurs at the first, second, or third object-goal
subtask in 102, 101, and 85 episodes, respectively.

\subsection{Post-Hoc Stage-Wise Shortest Paths}

For stage $k$ of episode $i$, the shortest-path denominator is recomputed from
the actual beginning of that trajectory segment:
\begin{equation}
d_{i,k}^{\star}
=
\min_{v\in\mathcal V(q_{i,k})}
d_{\mathrm{geo}}
\left(
x_{i,k}^{\mathrm{start}},v
\right),
\label{eq:supp_posthoc_distance}
\end{equation}
where $\mathcal V(q_{i,k})$ contains the valid navigable viewpoints of all
instances belonging to the requested object category.

Let $s_{i,k}\in\{0,1\}$ denote stage success and let $l_{i,k}$ be the executed
trajectory length. Stage-wise SPL is
\begin{equation}
\mathrm{SPL}_{i,k}
=
s_{i,k}
\frac{
d_{i,k}^{\star}
}{
\max(d_{i,k}^{\star},l_{i,k})
}.
\label{eq:supp_stage_spl}
\end{equation}

This post-hoc protocol does not require the agent to know future object goals
and avoids fixing later-stage shortest paths before the actual completion
location of the preceding goal is known.

\subsection{Conditional and Global Metrics}

Unexecuted later goals are excluded from conditional denominators. Setting
$s_{i,0}=1$, the conditional metrics are
\begin{equation}
\begin{aligned}
\mathrm{SR}^{\mathrm{cond}}_k
&=
\frac{
\sum_{i=1}^{N}s_{i,k}
}{
\sum_{i=1}^{N}s_{i,k-1}
},\\
\mathrm{SPL}^{\mathrm{cond}}_k
&=
\frac{
\sum_{i=1}^{N}
s_{i,k-1}\mathrm{SPL}_{i,k}
}{
\sum_{i=1}^{N}s_{i,k-1}
}.
\end{aligned}
\label{eq:supp_conditional_metrics}
\end{equation}

The corresponding global metrics use the complete episode set:
\begin{equation}
\begin{aligned}
\mathrm{SR}^{\mathrm{global}}_k
&=
\frac{1}{N}
\sum_{i=1}^{N}s_{i,k},\\
\mathrm{SPL}^{\mathrm{global}}_k
&=
\frac{1}{N}
\sum_{i=1}^{N}
s_{i,k-1}\mathrm{SPL}_{i,k}.
\end{aligned}
\label{eq:supp_global_metrics}
\end{equation}

The two views are related by
\begin{equation}
\begin{aligned}
\mathrm{SR}^{\mathrm{global}}_k
&=
\frac{\sum_i s_{i,k-1}}{N}
\mathrm{SR}^{\mathrm{cond}}_k,\\
\mathrm{SPL}^{\mathrm{global}}_k
&=
\frac{\sum_i s_{i,k-1}}{N}
\mathrm{SPL}^{\mathrm{cond}}_k.
\end{aligned}
\label{eq:supp_conditional_global_relation}
\end{equation}

Conditional metrics isolate performance among episodes that reach stage $k$,
whereas global metrics additionally preserve the cascade effect of earlier
failures. For a three-goal episode,
$\mathrm{SR}^{\mathrm{global}}_3$ is equal to the episode-level SR because
success at the third stage implies that all preceding goals have also been
completed.

\subsection{Cross-Target Memory and Perception Ablations}

The \textbf{w/o CT Memory} variant resets the accumulated geometric and
semantic voxel maps, stair connectivity, and related navigation memory when a
new object goal is issued. The simulator state and current agent pose are
retained, but the next goal is explored without cross-target map reuse. Thus,
this ablation removes the persistent cross-target memory rather than only
History-POI retrieval.

The \textbf{w/o Sem Gate} variant retains D-FINE-X detection, dense SED
features, History POIs, and the navigation policy, but disables the final
absolute SED similarity threshold of 0.15 used for target confirmation.
The perception comparison additionally reports a YOLOv7 configuration and the
complete D-FINE-X with semantic verification. The reported comparisons use the
same HM3D-MFMON task episodes.

\subsection{Global Results on All Episodes}

\begin{table}[t]
\centering
\caption{Stage-wise global performance on all 927
HM3D-MFMON episodes. All values are percentages.}
\label{tab:supp_all_global}
\scriptsize
\setlength{\tabcolsep}{2.2pt}
\resizebox{\columnwidth}{!}{
\begin{tabular}{lcccccc}
\toprule
\multirow{2}{*}{Method}
& \multicolumn{3}{c}{Global SR $\uparrow$}
& \multicolumn{3}{c}{Global SPL $\uparrow$} \\
\cmidrule(lr){2-4}
\cmidrule(lr){5-7}
& $k=1$ & $k=2$ & $k=3$
& $k=1$ & $k=2$ & $k=3$ \\
\midrule
OneMap
& 54.58 & 29.67 & 16.83
& 23.41 & 13.46 & 8.97 \\

w/o CT Memory
& \textbf{74.87} & 50.70 & 36.03
& \textbf{28.01} & 19.85 & 13.94 \\

LifelongCrossNav
& \textbf{74.87} & \textbf{51.56} & \textbf{36.57}
& \textbf{28.01} & \textbf{25.93} & \textbf{19.24} \\
\bottomrule
\end{tabular}}
\end{table}

\begin{table}[t]
\centering
\caption{Stage-wise global performance on the 288 CFR
episodes. All values are percentages.}
\label{tab:supp_cfr_global}
\scriptsize
\setlength{\tabcolsep}{2.2pt}
\resizebox{\columnwidth}{!}{
\begin{tabular}{lcccccc}
\toprule
\multirow{2}{*}{Method}
& \multicolumn{3}{c}{Global SR $\uparrow$}
& \multicolumn{3}{c}{Global SPL $\uparrow$} \\
\cmidrule(lr){2-4}
\cmidrule(lr){5-7}
& $k=1$ & $k=2$ & $k=3$
& $k=1$ & $k=2$ & $k=3$ \\
\midrule
OneMap
& 42.36 & 13.19 & 0.00
& 17.82 & 6.89 & 0.00 \\

w/o CT Memory
& \textbf{59.38} & 28.82 & 10.42
& \textbf{19.95} & 12.33 & 3.21 \\

LifelongCrossNav
& \textbf{59.38} & \textbf{29.17} & \textbf{12.50}
& \textbf{19.95} & \textbf{16.10} & \textbf{6.00} \\
\bottomrule
\end{tabular}}
\end{table}

The global results preserve the path-efficiency differences observed under
Conditional SPL while accounting for episodes that fail before reaching later
goals. LifelongCrossNav achieves higher second- and third-stage Global SPL than
w/o CT Memory, consistent with improved cross-target reuse of geometric and
semantic observations. The Global SPL values are computed from stage reach
counts and the reported Conditional SPL values; minor differences may arise
from rounding.

\subsection{Global Results on Cross-Floor-Required Episodes}

The CFR subset contains episodes whose first mandatory floor transition can
occur at any of the three object-goal stages. OneMap attains zero third-stage
Global SR, while LifelongCrossNav maintains nonzero third-stage Global SR and
SPL through explicit stair representation and cross-floor connectivity.
Compared with w/o CT Memory, the complete framework also achieves higher
later-stage Global SPL, indicating that reusing geometric and semantic memory
remains beneficial when floor transitions are required.

\end{document}